\documentclass[letterpaper]{article} 
\usepackage{aaai24}  
\usepackage{times}  
\usepackage{helvet}  
\usepackage{courier}  
\usepackage[hyphens]{url}  
\usepackage{natbib}  
\usepackage{caption} 
\usepackage{times}  
\usepackage{helvet}  
\usepackage{courier}  
\usepackage[hyphens]{url}  
\usepackage{graphicx} 
\usepackage{natbib}  
\usepackage{caption} 
\usepackage{times}
\usepackage{epsfig}
\usepackage{graphicx}
\usepackage{amsmath}
\usepackage{amssymb}
\usepackage{amsmath}
\usepackage{amssymb}
\usepackage{booktabs}
\usepackage{multirow}
\usepackage{makebox}
\usepackage{array}
\usepackage{subcaption}
\usepackage[dvipsnames]{xcolor}
\usepackage{amssymb}
\usepackage[title]{appendix}
\usepackage{xspace}
\usepackage{mathtools}
\usepackage{bbding}
\usepackage{makecell}

\usepackage{pifont}
\usepackage{url}
\usepackage{nicefrac}
\newcommand{\cmark}{\ding{51}}%
\newcommand{\xmark}{\text{\ding{55}}}

\newcommand{\pos}[1]{#1_{\omega}}
\definecolor{mygray}{gray}{.9}
\usepackage{colortbl}

\usepackage{algorithm}
\usepackage{algorithmic}

\usepackage{newfloat}
\usepackage{listings}
\DeclareCaptionStyle{ruled}{labelfont=normalfont,labelsep=colon,strut=off} 
\floatstyle{ruled}
\newfloat{listing}{tb}{lst}{}
\floatname{listing}{Listing}
\title{Self-Supervised Likelihood Estimation with Energy Guidance for Anomaly Segmentation in Urban Scenes}
\author{
    Yuanpeng Tu\textsuperscript{\rm 1}\equalcontrib, 
    Yuxi Li\textsuperscript{\rm 2}\equalcontrib,
    Boshen Zhang\textsuperscript{\rm 2}\equalcontrib,
    Liang Liu\textsuperscript{\rm 2},\\
    Jiangning Zhang\textsuperscript{\rm 2},
    Yabiao Wang\textsuperscript{\rm 2 $\dagger$ },
    Cai Rong Zhao\textsuperscript{\rm 1}\thanks{Corresponding author.}
}
\affiliations{
    \textsuperscript{\rm 1} Dept. of Electronic and Information Engineering, Tongji Univeristy, Shanghai\\
    \textsuperscript{\rm 2} YouTu Lab, Tencent, Shanghai \\
	{ \{2030809, zhaocairong\}@tongji.edu.cn}\\
	{ \{yukiyxli, boshenzhang, leoneliu, vtzhang, caseywang\}@tencent.com}}
\usepackage{bibentry}

\begin{document}

\maketitle

\begin{abstract}
Robust autonomous driving requires agents to accurately identify unexpected areas (anomalies) in urban scenes. To this end, some critical issues remain open: how to design advisable metric to measure anomalies, and how to properly generate training samples of anomaly data? Classical effort in anomaly detection usually resorts to pixel-wise uncertainty or sample synthesis, which ignores the contextual information and sometimes requires auxiliary data with fine-grained annotations. On the contrary, in this paper, we exploit the strong context-dependent nature of the segmentation task and design an energy-guided self-supervised framework for anomaly segmentation, which optimizes an anomaly head by maximizing the likelihood of self-generated anomaly pixels. For this purpose, we design two estimators to model anomaly likelihood, one is a task-agnostic binary estimator and the other depicts the likelihood as residual of task-oriented joint energy. Based on the proposed estimators, we devise an adaptive self-supervised training framework, which exploits the contextual reliance and estimated likelihood to refine mask annotations in anomaly areas. We conduct extensive experiments on challenging Fishyscapes and Road Anomaly benchmarks, demonstrating that without any auxiliary data or synthetic models, our method can still achieve comparable performance to supervised competitors. Code is available at \url{https://github.com/yuanpengtu/SLEEG}.

\end{abstract}

\section{Introduction}
\label{sec:intro}
Recent studies in semantic segmentation have achieved significant advances on close-set benchmarks of urban scenarios~\cite{cordts2016cityscapes}. However, when it comes to deployment in the wild, it is necessary to enable segmentation models with the ability of anomaly detection, i.e., to ensure segmentation networks can identify some Out-of-Distribution (OoD) objects which did not appear in its training set. 

Essentially, the key of segmentation with anomalies lies in two aspects: \emph{Firstly,} the anomaly score should be carefully designed to effectively differentiate anomaly pixels and normal ones. \emph{Second,} other than normal data from segmentation datasets, extra anomaly data is critical to explicitly identify which pixel belongs to anomaly areas. To address these issues, a fresh wave of approaches in semantic segmentation are proposed recently. To measure the likelihood of anomalies, some methods take insight from uncertainty estimation and devise a series of proxy tasks~\cite{MSP,bayesian2018,diric2018,grcic2022densehybrid,tian2021pixel}. However, these approaches usually design coupled objectives for both segmentation and anomaly detection and require retraining of segmentation models, which might degrade the performance on both tasks~\cite{cvpr2022_survey}. On the other hand, to generate training samples with anomaly pixels, outlier exposure is widely adopted~\cite{bevandic2019simultaneous,grcic2022densehybrid,chan2021entropy} by training models with auxiliary data, while in a practical sense, these methods increase the cost due to additional labeling requirements, the adopted auxiliary data is also not guaranteed to be consistent with realistic scenes. There are also approaches leveraging an extra reconstruction model and taking the reconstruction error as anomaly parts~\cite{xia2020synthesize,synboost2021,vojir2021road}, which affects the efficiency and the detection results highly rely on reconstruction quality. 

\begin{figure}
    \centering 
    \includegraphics[width=0.48\textwidth]{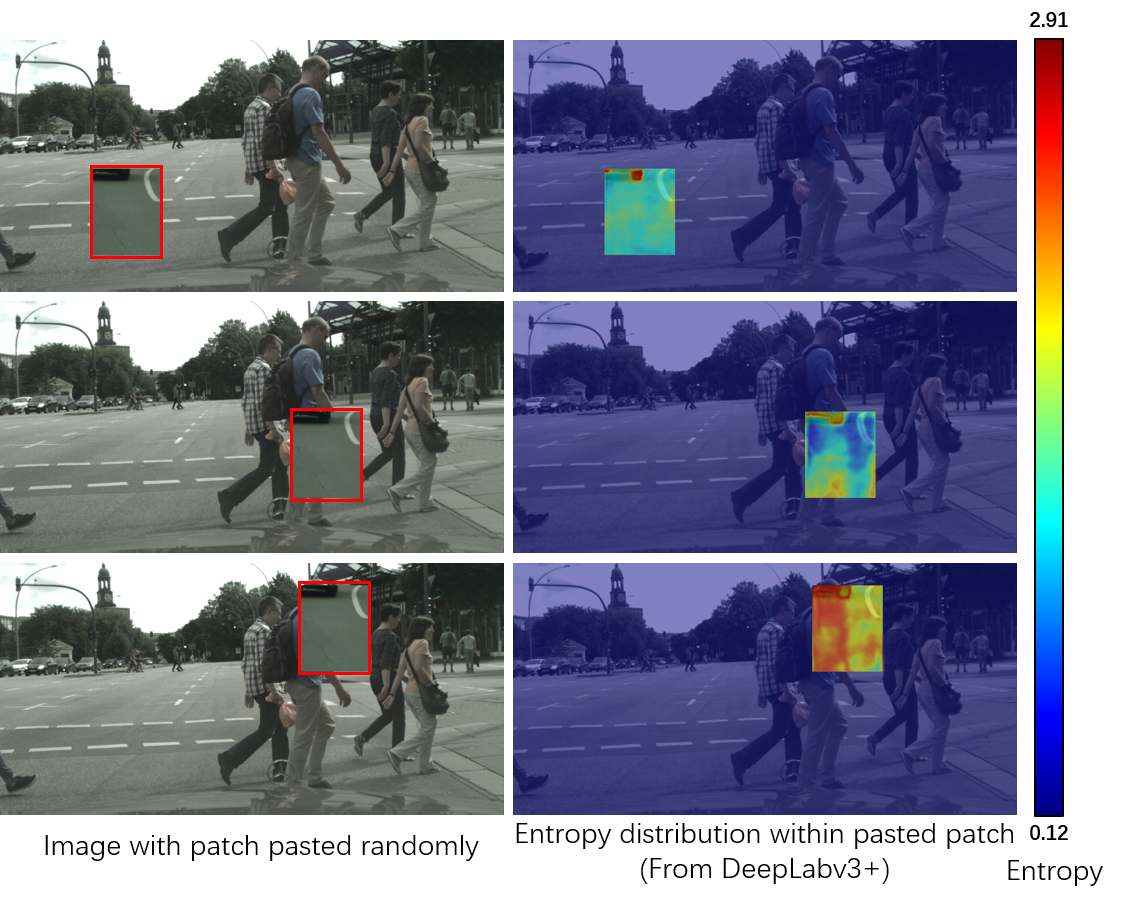}
    \vspace{-2em}
    \caption{Illustration of contextual reliance in anomaly segmentation tasks. The \textbf{left} column shows an image pasted with a random patch at different position. The \textbf{right} column illustrates the corresponding entropy distribution from segmentation results of DeepLab segmentation model~\cite{chen2018encoder}. Different pasted positions result in various uncertainty distribution within the patch.}
    \label{fig:motivation}
    \vspace{-7mm}
\end{figure}

In summary, most previous efforts are originally designed to capture anomaly samples in classification. Nevertheless, semantic segmentation differs since the segmentation results inherently rely on spatial context. As shown in Fig.~\ref{fig:motivation}, given a pretrained segmentation model~\cite{chen2018encoder}, the same patch yield different semantic uncertainty (measured as entropy of categorical distribution) when it is placed under a different context, even though the patch is filled with normal pixels. The empirical observation inspires that \textbf{we can automatically synthesize anomalies from normal pixels via a self-supervised copy-and-paste manner guided by spatial context.} Such self-supervised paradigm can (1) avoid the cost of explicitly annotating anomalies and (2) ensure the quality of generated anomalies by referring to their context.
Therefore, we propose a new framework termed as \textbf{S}elf-supervised \textbf{L}ikelihood \textbf{E}stimation with \textbf{E}nergy \textbf{G}uidance (\textbf{SLEEG}), which extends off-the-shelf segmentation models to anomaly detectors with the guidance of energy model~\cite{lecun2006tutorial} while avoiding the overhead of labeling anomaly data. 

The SLEEG framework is designed in a self-teaching paradigm, to properly depict the anomaly area, we propose two anomaly estimators based on the joint distribution of content and anomalies. The first is formulated as a simple task-agnostic classifier to differentiate anomaly and normal pixels. The other is a task-oriented estimator and can be naturally regarded as residual estimation of classic joint-energy model (JEM)~\cite{grathwohl2019your}, with proper design of loss function, this estimator can be optimized through a dynamic energy-guided margin. Next, based on these estimators, we design an adaptive refinement mechanism to provide dynamic anomaly samples under different contexts, which is guided by anomaly likelihood and contextual information of each pixel.

We implemented SLEEG with different segmentation models and evaluate the performance on benchmarks of Fishyscapes~\cite{blum2019fishyscapes} and Road Anomaly~\cite{imgr}. Experiments results show that SLEEG can bring consistent improvement over different baselines by training with only normal data from Cityscapes~\cite{cordts2016cityscapes}. Further, compared with other state-of-the-art anomaly segmentation methods, SLEEG achieves competitive results \textbf{without training on labeled anomaly data or updating parameters of segmentation networks.} In summary, the contributions of this paper can be listed as:

\begin{itemize}
    \item We propose SLEEG, a self-supervised framework for anomaly segmentation in a copy-and-paste manner. In the framework, we design two decoupled likelihood estimators from an energy model, a task-agnostic estimator for discriminative learning and a task-oriented estimator for residual learning of joint-energy.
    \item Based on the proposed energy-guided estimators, we propose a dynamic mask refinement mechanism by applying likelihood-guided pixel separation to help extract more informative anomaly areas for training.
    \item Without training on labeled auxiliary data or updating segmentation parameters, our SLEEG achieves competitive performance on both Fishyscapes~\cite{blum2019fishyscapes} and Road Anomaly~\cite{imgr} benchmarks.
\end{itemize}

\section{Related Work}
\begin{figure*}[!t]
 \centering
 \begin{minipage}{0.9\textwidth}
    \centering
    \includegraphics[width=\textwidth]{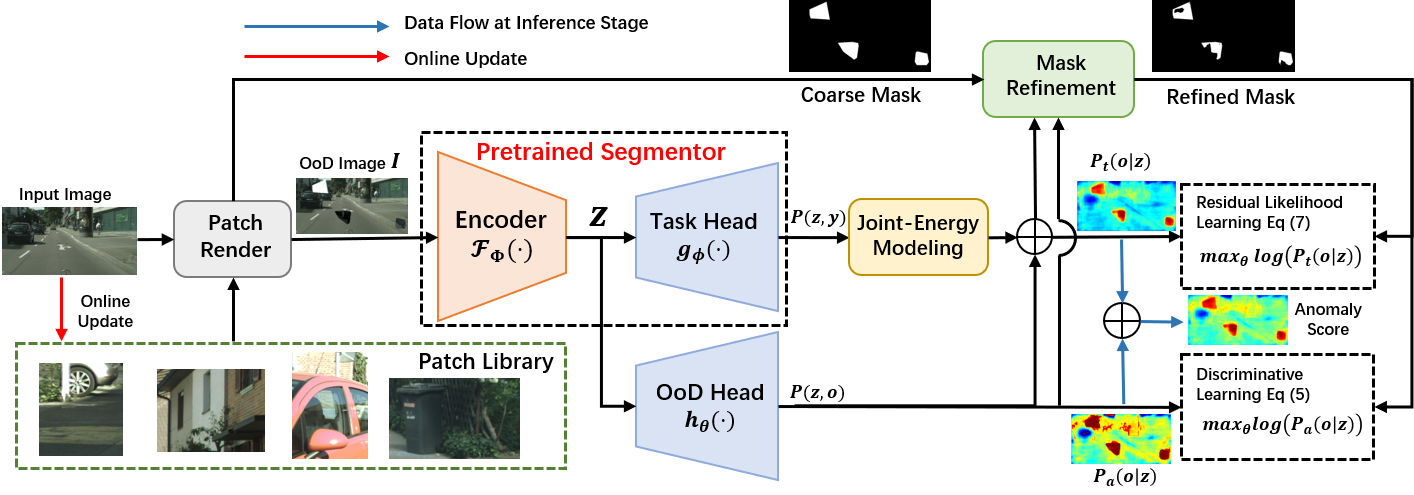}
\end{minipage}

\vspace{-1em}
  \caption
    {
        Illustration of proposed SLEEG framework, an OoD head is extended and trained in a self-supervised manner to enable a pretrained segmentation model with anomaly detection ability.   
      }
  \label{fig:framework}
  \vspace{-6mm}
 \end{figure*}
 
\subsection{Anomaly Segmentation}
Some of previous effort extends technique from anomaly detection (e.g. uncertainty estimation and outlier exposure) in classification into segmentation tasks to help identify anomaly pixels, recently there also appears new paradigm which exploits reconstruction error to highlight anomalies.

\noindent\textbf{Uncertainty Estimation.}
Similar to image-level anomaly detection approaches, early uncertainty based methods~\cite{hendrycks2016baseline,lee2017training} focused on measuring anomaly scores of each pixel with maximum softmax probability produced by the softmax classifier since the model tends to output uniform prediction for unseen semantics. However, they are prone to misclassify pixels of tail classes as anomalies, since the same threshold is set for all pixels regardless of the class-wise distribution discrepancy. To address this issue, Jung \emph{et al.}~\cite{jung2021standardized} proposed standardized max logit (SML), which normalized the distribution of max logit from seen classes and mitigates the influence of anomalous boundary pixels with neighboring-pixel iterative refinement, bringing considerable improvement. Recent methods try to enhance ability of distinguishing hard samples from anomalous ones by re-training the classifiers with anomaly objectives. However, these approaches generally suffer from accuracy decrease on seen categories~\cite{cvpr2022_survey}. 

\noindent\textbf{Outlier Exposure.}
Recent promising methods~\cite{bevandic2018discriminative, chan2021entropy,di2021pixel, bevandic2019simultaneous,hendrycks2018deep} are intuitive, which utilize labeled samples from non-overlapped classes of an external auxiliary dataset as anomalies to help models differentiate unexpected instances against normal pixels. Hendrycks et al.~\cite{hendrycks2018deep} made the first attempt by forcing model to predict uniform distribution of anomaly detection. ~\cite{chan2021entropy} and ~\cite{di2021pixel} leveraged instance masks from COCO~\cite{lin2014microsoft} and void category from Cityscapes~\cite{cordts2016cityscapes} respectively to make model generalize to unexpected objects. However, they usually require re-training of the model and suffer from potential degradation in accuracy of in-distribution recognition by maximizing uncertainty on anomalies. Besides, these strategies require fine-grained annotation (e.g. instance mask or bounding boxes) of anomaly objects, inevitably increasing the cost of labeling. Finally, utilizing specific external datasets as outliers may lead the anomaly detectors biased toward different domain, resulting in accuracy deterioration in real-world scenes.   

\noindent\textbf{Image Reconstruction.}
Methods based on image reconstruction~\cite{lis2019detecting, xia2020synthesize} usually employ generative adversarial networks (GANs)~\cite{creswell2018generative} to fit the distribution of normal pixels, re-synthesize images conditioned on predicted segmentation results and localize the discrepancy between original images and reconstructed ones as anomalous objects. Nevertheless, these approaches usually heavily rely on the accurate segmentation maps and performance of reconstruction model, while it is still difficult for the segmentation models to distinguish hard in-distribution pixels and anomalous ones. On the other hand, their performance can be affected by the artifacts generated by GANs as well. Finally, they also suffer from time-consuming serialized training and inference processes of the reconstruction networks, making them hard to be applied in real-time scenarios~\cite{cvpr2022_survey}. 

\subsection{Energy Based Modeling}
There is also a special series of methods applying energy function~\cite{lecun2006tutorial} to depict probabilities of anomaly data. These approaches apply a free-energy function~\cite{lecun2006tutorial} as anomaly indicators and focus on minimizing energy for normal instances while maximizing energy for unseen outlier samples. Afterwards, the energy value is taken as measurement to predict anomaly probability of samples. Previous energy-based models generally employ Markov Chain Monte Carlo as the partition function to estimate energy score whereas high-quality samples cannot be generated in this manner. To address this issue, ~\cite{tian2021pixel} takes the insight from absenting learning, and utilizes the joint-energy~\cite{grathwohl2019your} with additional smooth terms to help switch between normal classification task and anomaly detection, while still requiring outlier exposure strategy with fine-grained annotation.

\section{Methodology}

In this section we introduce the detail of SLEEG framework, which is depicted as Fig.~\ref{fig:framework}. Given an image $\mathcal{I} \in \mathbb{R}^{3\times H\times W}$ with $H, W$ indicating its spatial resolution, its spatial coordinate set is defined as $\Omega$, we associate the pixel $\pos{x}$ at each coordinate $\omega \in \Omega$ with a triplet variable $(\pos{z}, \pos{y}, \pos{o})$, where $\pos{z} \in \mathbb{R}^D$ represents the encoded feature of dimension $D$, $\pos{y} \in \{0, 1,\cdots, K-1\}$ denotes predicted categorical labels over $K$ close-set semantic classes, and $\pos{o} \in \{0,1\}$ is a binary indicator to denote whether $\pos{x}$ belongs to anomaly. Our approach follows the classical Encoder-Decoder meta-architecture in semantic segmentation, where an encoder $\mathcal{F}_{\Phi}(\cdot) : \mathbb{R}^{3\times H\times W} \rightarrow \mathbb{R}^{D\times H\times W} $ extracts deep features and a segmentation decoder $g_{\phi}(\cdot) : \mathbb{R}^{D} \rightarrow \mathbb{R}^{K}$ is responsible to predict categorical distribution, both $\Phi,\phi$ can be pretrained parameters from off-the-shelf segmentation models. Similarly, we further extend a learnable anomaly decoder $h_{\theta}(\cdot): \mathbb{R}^{D} \rightarrow \mathbb{R}^{2}$ (termed as OoD head in Fig.~\ref{fig:framework}) from original segmentation model, the output of which is utilized to derive two different anomaly estimators. The learnable parameter $\theta$ is trained in a self-supervised pipeline by maximizing the estimator scores in pseudo anomaly areas.

\subsection{Anomaly Estimators for Likelihood Maximization}\label{sec:likelihood}

In anomaly detection, our goal is to accurately model the anomaly likelihood given input data $p(\pos{o}|\pos{z})$. To this end, we resort to the Bayes Rule to derive the conditional probability, however, since the feature encoding $\pos{z}$ is jointly modeled with both semantic $\pos{y}$ and anomaly $\pos{o}$, the likelihood probability can be derived with different marginalization
\begin{equation}\label{eq:marginal}
    p(\pos{o}|\pos{z}) = \frac{p(\pos{z}, \pos{o})}{\sum_{\pos{o}=0}^1{p(\pos{z}, \pos{o})}} =  \frac{p(\pos{z}, \pos{o})}{\sum_{\pos{y}=1}^K{p(\pos{z}, \pos{y})}}
\end{equation}
This converts the estimation from conditional distribution $p(\pos{o}|\pos{z})$ to the joint distribution of $p(\pos{z}, \pos{o})$ and $p(\pos{z}, \pos{y})$. To analytically estimate the joint distribution, we recap the energy-based model~\cite{lecun2006tutorial,grathwohl2019your} by reinterpreting the decoder $g_{\phi}(\cdot)$ as an energy function, which estimates the joint distribution $p(\pos{z},\pos{y})$
\begin{equation}\label{eq:zy}
    p(\pos{z},\pos{y};\phi) = \frac{1}{\mathcal{T}(\phi)}\exp(g_{\phi}(\pos{z}))[\pos{y}]
\end{equation}
where $[\pos{y}]$ is the $\pos{y}$-th index of output categorical vector, and $\mathcal{T}(\phi)=\int_{\pos{z}}\sum_{\pos{y}}{\exp(g_{\phi}(\pos{z}))[\pos{y}]}d\pos{z}$ is an unknown normalization factor. 
Similarly, the learnable anomaly decoder $h_{\theta}(\cdot)$ can be regarded as an energy function to estimate the distribution of $p(\pos{z},\pos{o})$
\begin{equation}\label{eq:zo}
    p(\pos{z},\pos{o};\theta) = \frac{1}{\Gamma(\theta)}\exp(h_{\theta}(\pos{z}))[\pos{o}]
\end{equation}
where $\Gamma(\theta)$ is another constant factor similar to $\mathcal{T}(\phi)$. By inserting Eq.~(\ref{eq:zo}) into Eq.~(\ref{eq:marginal}), we can derive analytical representation of $p(\pos{o}|\pos{z})$ which only focuses on anomalies regardless of segmentation tasks. On the other hand, when inserting both Eq~.(\ref{eq:zo}) and Eq~.(\ref{eq:zy}) into Eq.~(\ref{eq:marginal}), we essentially take semantic distribution $p(\pos{z}, \pos{y})$ into account for estimation. Hence we obtain two different anomaly estimators.

\noindent\textbf{Task Agnostic Estimator (TAE).} By taking the anomaly segmentation as a pixel-wise binary classification problem, we can easily derive the likelihood from Eq.~(\ref{eq:zo}) and Eq.~(\ref{eq:marginal}) with the normalization factor $\Gamma(\theta)$ eliminated
\begin{equation}\label{eq:tae}
   p_a(\pos{o}|\pos{z};\theta) = \frac{\exp(h_{\theta}(\pos{z}))[\pos{o}]}{\sum_{\pos{o} \in \{0, 1\}}\exp(h_{\theta}(\pos{z}))[\pos{o}]}
\end{equation}
Since Eq.~(\ref{eq:tae}) is a normalized probability function, we can optimize this estimator via simple cross-entropy loss
\begin{align}\label{eq:loss1}
\mathcal{L}_{a}(\theta) = & -E_{\pos{x}\in \mathcal{S}_{ood}}\left[\log{p_a(\pos{o}=1|\pos{z};\theta)}\right] \nonumber\\
                            & - E_{\pos{x}\in \mathcal{S}_{id}}\left[\log{p_a(\pos{o}=0|\pos{z};\theta)}\right]
\end{align}
where $\mathcal{S}_{ood}$ denote the set of anomaly pixels and $\mathcal{S}_{id}$ is the set of normal pixels.

\noindent\textbf{Task Oriented Residual Estimator (TORE)}. When taking joint probability $p(\pos{z}, \pos{y})$ from Eq.~(\ref{eq:zy}) for marginalization in Eq.~(\ref{eq:marginal}), the estimated likelihood $p_t(\pos{o}|\pos{z};\theta)$ is coupled with constants $\Gamma(\theta)$ and $\mathcal{T}(\phi)$, which is intractable, hence we transform the likelihood into logarithmic form
\begin{align}\label{eq:tore}
    \log{p_t(\pos{o}|\pos{z};\theta)} & = h_{\theta}(\pos{z})[\pos{o}] + \text{JEM}(\pos{z}) + C(\phi, \theta) \nonumber \\
     \text{JEM}(\pos{z}) & = - \log{\sum_{\pos{y}}\exp(g_{\phi}(\pos{z}))[\pos{y}]}
\end{align}
where $C(\phi, \theta)=\log{(\mathcal{T}(\phi)/\Gamma(\theta))}$ is a constant w.r.t $\pos{z}$. Note that the second term in Eq.~(\ref{eq:tore}) is exactly the negative joint-energy model (JEM)~\cite{grathwohl2019your}, which can be regarded as ``coarse'' estimation of uncertainty, therefore Eq.~(\ref{eq:tore}) essentially takes decoder $h_{\theta}$ to estimate the residual of JEM score, hence to ``refine'' the anomaly area. To optimize the estimator while handling intractable factor $C(\phi, \theta)$, we exploit a margin loss to compare the likelihood of anomaly and normal pixels and eliminate additive constant $C(\phi, \theta)$
\begin{align}\label{eq:loss2}
    \mathcal{L}_{o}(\theta) = \{E_{\pos{x}\in\mathcal{S}_{id}}[\log{p_t(\pos{o}=1|\pos{z};\theta)}] - \nonumber\\ E_{x_{\omega'}\in\mathcal{S}_{ood}}[\log{p_t(o_{\omega'}=1|z_{\omega'};\theta)}] + \gamma\}_{+}
\end{align}
where $\gamma$ is a hyperparameter to control the margin, $\{\cdot\}_{+}$ indicates truncating the value to $0$ if it is negative. 

\noindent\textbf{Difference from other energy-based methods.} Note that the loss term in Eq.~(\ref{eq:loss2}) can be reformulated as\footnote{For simplicity, we ignore the index symbol $[\pos{o}=1]$ here.}
\begin{align}\label{eq:dynamic}
     \mathcal{L}_{o}(\theta) & = \{E_{\pos{x}\in\mathcal{S}_{id}}[h_{\theta}(\pos{z})] - E_{x_{\omega'}\in\mathcal{S}_{ood}}[h_{\theta}(z_{\omega'})] + \widehat{\gamma}\}_{+} \nonumber\\
     \widehat{\gamma} = \gamma & + E_{\pos{x}\in\mathcal{S}_{id}}[\text{JEM}(\pos{z})] - E_{x_{\omega'}\in\mathcal{S}_{ood}}[\text{JEM}(z_{\omega'})]
\end{align}
therefore, our TORE is different from other methods that directly update classifiers to optimize JEM~\cite{grcic2022densehybrid,tian2021pixel}, instead it takes estimated joint-energy to dynamically control the margin $\widehat{\gamma}$ of anomaly confidence between anomaly and normal pixels. This dynamic margin $\widehat{\gamma}$ is also verified more helpful than a static margin $\gamma$ in the ablation study.

Finally, the predicted anomaly score can be expressed as the combination of both estimators
\begin{equation}\label{eq:inference}
    \mathcal{A}(\pos{x}) = \log{p_a(\pos{o}|\pos{z};\theta^*)} + \lambda\log{p_t(\pos{o}|\pos{z};\theta^*)}
\end{equation}
where $\lambda$ is a balance factor and $\theta^*$ denotes the parameters of OoD head after optimization. In Eq.~(\ref{eq:inference}) we omit the constant term $C(\phi, \theta)$ since it does not affect the relative order of different pixels.


\subsection{Self-supervised Training with Refined Patch}\label{sec:training}
In this section we describe our self-training pipeline, which is illustrated in Fig.~\ref{fig:framework}. Different from classical outlier exposure~\cite{Denseoutlier2021,synboost2021}, we aim at generating training samples with both anomaly pixels $\mathcal{S}_{ood}$ and normal area $\mathcal{S}_{id}$ in a self-supervised manner without manual mask annotation.

\noindent\textbf{Anomaly Patch Rendering.} Intuitively, due to the contextual reliance of segmentation network, a random patch can be an anomaly part once placed under somewhat unnatural pattern, even if the patch is cropped from a normal image in training scenes. With this consideration, we design a dynamic copy-and-paste strategy to generate pseudo anomaly samples (as shown in Fig.~\ref{fig:framework}). In detail, for each input image, we first randomly crop $N$ rectangle patches from other images as candidates. 
To ensure the geometric diversity of anomalies, we extract the Harris corner points~\cite{harris1988combined} within each candidate and crop the minimum polygon containing all these points as anomaly patch, finally, all cropped polygons are randomly pasted on input image.

\begin{figure}[!t]
 \centering
 \begin{minipage}{0.47\textwidth}
    \centering
    \includegraphics[width=0.92\textwidth]{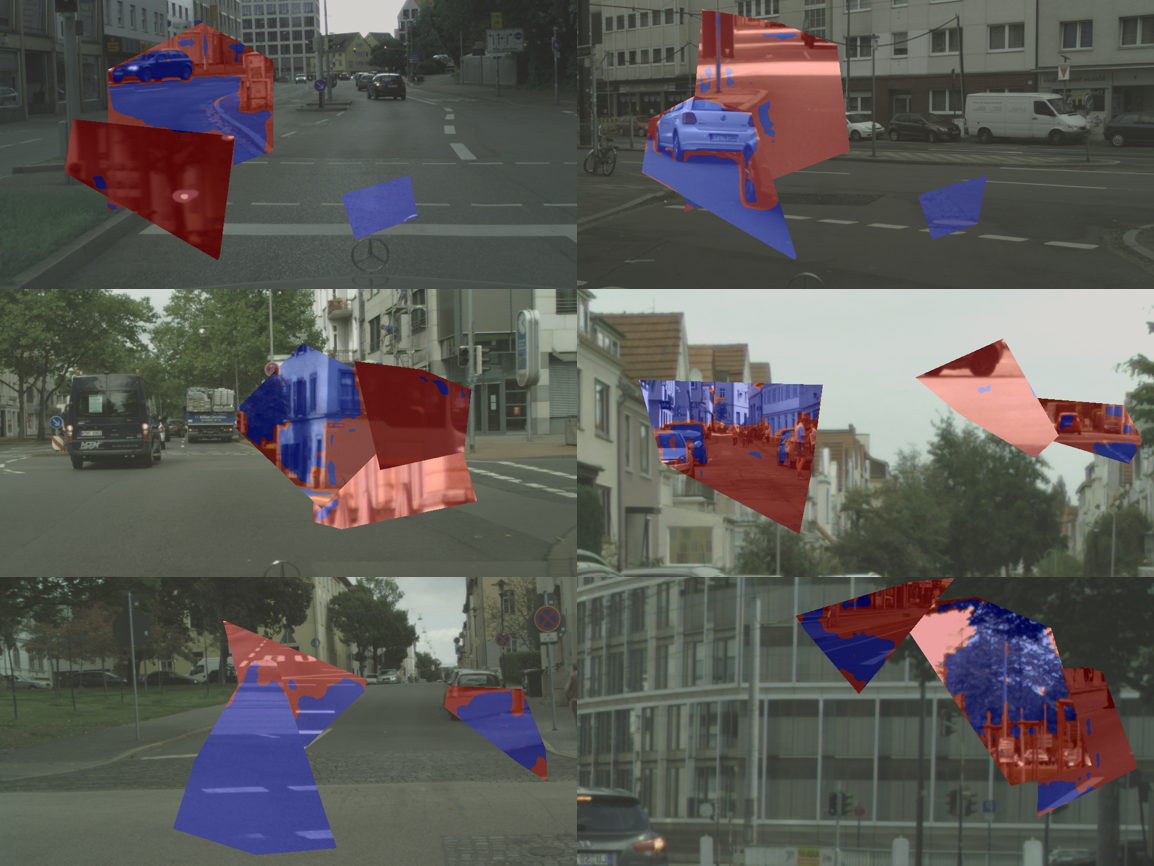}
\end{minipage}

\vspace{-1em}
  \caption
    {
        Visualization of generated patches with random shapes on training set of Cityscapes. Area with \textcolor{red}{\textbf{red}} mask denotes anomaly pixels after mask refinement, the \textcolor{blue}{\textbf{blue}} area represents the ignored pixels from pasted patches.
      }
  \label{fig:energy_visualize}
  \vspace{-5mm}
 \end{figure}

\noindent\textbf{Adaptive Mask Refinement.} 
As the patches are randomly pasted, they can still contain objects which fits the context well. Taking these pixels as anomaly will reversely hinder the detection results. Therefore, to fully leverage the contextual information, we take the estimated anomaly likelihood $\mathcal{A}(\pos{x})$ as guidance to measure the inconsistency between context and pasted patches and further refine the pasted polygons. Formally, we define the coordinates set of the $i$-th pasted polygon area as $\Omega^p_i$, and aim at finding a threshold $\eta^*$ to separate out pixels more likely to be anomalies
\begin{equation}
    \mathcal{S}_{ood} = \{\pos{x}| \omega \in \cup_{i=1}^{N}\Omega^p_i \wedge \mathcal{A}(\pos{x}) \geq \eta^* \}
\end{equation}
To properly refine the pasted patch, we design the threshold $\eta^*$ in a dynamic manner such that pixels within each separated group share similar anomaly likelihood. Therefore we search the threshold by minimizing the variance of anomaly scores within each pixel group in pasted area
\begin{align}\label{eq:otsu}
    & \eta^* = \arg\min_{\eta} Var_{\pos{x}\in\mathcal{S}_{ood}}[\mathcal{A}(\pos{x})] + Var_{\pos{x}\notin{\mathcal{S}_{ood}}}[\mathcal{A}(\pos{x})] \nonumber\\
    & \textbf{s.t. } \hspace{2mm} \omega \in \cup_{i=1}^{N} \Omega^p_i \quad \min_{\omega}{\mathcal{A}(\pos{x})} \leq \eta \leq \max_{\omega}{\mathcal{A}(\pos{x})}
\end{align}
The problem in Eq.~(\ref{eq:otsu}) is solved in a grid scanning manner, which is equivalent to finding the maximum gap between mean values of groups ~\cite{otsu1979threshold}. After the refinement, the pasted areas not attributed to $\mathcal{S}_{ood}$ is labeled as ignored and will not be used for loss computation. On the other hand, we take all pixels outside the pasted area as normal set $\mathcal{S}_{id}$. Fig.~\ref{fig:energy_visualize} shows some examples after our refinement, it can be clearly observed that some part fitting the context well is automatically filtered out and other areas that are more contradictory to context are left for training. With the set separation above, we can reversely apply Eq.~(\ref{eq:loss1}) and Eq.~(\ref{eq:loss2}) to train the OoD head in a self-supervised paradigm.

\begin{table*}[ht!]
\small
  \centering
  \begin{tabular}{l|c|c|cc|cc}
    \toprule
    \multirow{2}*{Method} &  \multirow{2}{*}{\shortstack{OoD Data}}& \multirow{2}{*}{\shortstack{Re-training}} & \multicolumn{2}{c}{FS LAF} & \multicolumn{2}{|c}{FS Static} \\
    \cline{4-7} & & & $\rm{FPR_{95}}$ $\downarrow$ & AP $\uparrow$ & $\rm{FPR_{95}}$ $\downarrow$ & AP $\uparrow$\\
    \midrule
    \midrule
    {Synboost}~\cite{synboost2021} & \cmark &\xmark & 15.79 & 43.22 & 18.75 & 72.59  \\
    {Density - Logistic Regression}~\cite{blum2021fishyscapes} & \cmark &\cmark & 24.36 & 4.65 & 13.39 & 57.16  \\
    {Bayesian Deeplab}~\cite{bayesian2018} & \xmark &\cmark & 38.46 & 9.81 & 15.50 & 48.70  \\
    {OoD Training - Void Class}~\cite{blum2021fishyscapes}& \cmark &\cmark & 22.11 & 10.29 & 19.40 & 45.00  \\
    {Discriminative Outlier Detection Head}~\cite{Denseoutlier2021} & \cmark &\cmark & 19.02 & 31.31 & \textbf{0.29} & \textbf{96.76}  \\
    {Dirichlet Deeplab}~\cite{diric2018}& \cmark &\cmark & 47.43 & 34.28 & 84.60 & 31.30 \\
    
    {DenseHybrid}$\dagger$ ~\cite{grcic2022densehybrid} & \cmark &\cmark & \underline{6.18} & {43.90} & 5.51 & 72.27 \\
     {PEBAL}$\dagger$ ~\cite{di2021pixel} & \cmark &\cmark & {7.58} & \underline{44.17} & {1.73} & {92.38} \\

    {CoroCL}$\dagger$~\cite{liu2023residual} & \cmark  & \cmark & \textbf{2.27}  & \textbf{53.99}  & \underline{0.52}  &\underline{95.96}   \\

    \midrule
    {MSP}~\cite{MSP}& \xmark &\xmark & 44.85 & 1.77 & 39.83 & 12.88 \\
    {Entropy}~\cite{MSP}& \xmark &\xmark & 44.83 & 2.93 & 39.75 & 15.41 \\
    {Density - Single-layer NLL}~\cite{blum2021fishyscapes}& \xmark &\xmark & 32.90 & 3.01 & 21.29 & 40.86  \\
    {kNN Embedding - density}~\cite{blum2021fishyscapes} & \xmark &\xmark & 30.02 & 3.55 & 20.25 & 44.03  \\
    {Density - Minimum NLL}~\cite{blum2021fishyscapes}& \xmark &\xmark & 47.15 & 4.25 & 17.43 & 62.14  \\
    {Image Resynthesis}~\cite{imgr}& \xmark &\xmark & 48.05 & 5.70 & 27.13 & 29.60  \\
    {SML}~\cite{jung2021standardized} & \xmark &\xmark & 21.52 & 31.05 & 19.64 & 53.11  \\
    {GMMSeg}~\cite{liang2022gmmseg} & \xmark &\xmark & \underline{6.61} & \underline{55.63} & \underline{15.96} & \textbf{76.02}\\
    {FED-U}~\cite{gudovskiy2023concurrent} & \xmark &\xmark & {11.38} & {20.45} & {21.58} & {67.80}\\
    {FED-C}~\cite{gudovskiy2023concurrent} & \xmark &\xmark & \textbf{5.20} & {50.15} & {31.97} & {61.06}\\
    \rowcolor{mygray}
    {SLEEG (ours)} & \xmark &\xmark & {6.69} & \textbf{59.66} & \textbf{10.49} & \underline{68.93}\\
    \bottomrule
  \end{tabular}
  \vspace{-.8em}
  \caption{Comparison with previous methods on the FS test set. ``OoD Data'' indicates training with additional labeled anomaly data. ``Re-training'' means updating parameters of segmentation network. 
   $\dagger$ indicates training with WideResNet38. \textbf{Bold} values and \underline{underlined} values represent the best and second best results (Comparison is conducted within each group).}
  \label{tab:FS_leaderboard}
  \vspace{-4mm}
\end{table*}

 \begin{table*}[ht!]
 \scriptsize
  \centering
  \setlength{\tabcolsep}{1.85mm}{
  \begin{tabular}{l|c|ccc|ccc|ccc}
    \toprule
    \multirow{2}*{Method} & \multirow{2}{*}{\makecell{OoD \\ Data}} & \multicolumn{3}{c}{FS LAF Validation} & \multicolumn{3}{|c}{FS Static Validation}& \multicolumn{3}{|c}{Road Anomaly Test}\\
    \cline{3-11} & & $\rm{FPR_{95}}$ $\downarrow$ & AUROC $\uparrow$ & AP $\uparrow$ & $\rm{FPR_{95}}$ $\downarrow$ & AUROC $\uparrow$ & AP $\uparrow$& $\rm{FPR_{95}}$ $\downarrow$ & AUROC $\uparrow$ & AP $\uparrow$ \\
    \midrule
    \midrule
    {Synboost}~\cite{synboost2021} & \cmark & 34.47  & 94.89  & 40.99 & 47.71 & 92.03 & 48.44 & 59.72 & 85.23 & {41.83} \\
    {DenseHybrid}$\dagger$ ~\cite{grcic2022densehybrid} & \cmark & {6.10} & - & {63.80}  & 4.90 & - & 60.00 & - & -  & - \\
    {PEBAL}$\dagger$ ~\cite{tian2021pixel} & \cmark &  {4.76}  & {98.96}  & 58.81 & \textbf{1.52} & \textbf{99.61} & \textbf{92.08} & \textbf{44.58} & \textbf{87.63} & \textbf{45.10} \\
    {CoroCL}$\dagger$~\cite{liu2023residual} & \cmark & \textbf{0.85} & \textbf{99.73}  &\textbf{92.46}  &2.52  &99.39 &70.61 & -   &-   &- \\
    
    \midrule
    {MSP}~\cite{MSP} & \xmark & 45.63 & 86.99 & 6.02 & 34.10 & 88.94 & 14.24 & 68.44 & 73.76 & 20.59\\
    {MaxLogit}~\cite{streethazards2019} & \xmark & 38.13 & 92.00 & 18.77 & 28.50 & 92.80 & 27.99 & 64.85 & 77.97 & 24.44\\
    {SynthCP}~\cite{xia2020synthesize} & \xmark & 45.95 & 88.34 & 6.54 & 34.02 & 89.90 & 23.22 & 64.69 & 76.08 & 24.87 \\
    {SML}~\cite{jung2021standardized} & \xmark & 14.53 & 96.88 & 36.55 & 16.75 & 96.69 & 48.67 & 49.74 & 81.96 & 25.82 \\
    {LDN\_BIN~\cite{Denseoutlier2021}} & \xmark & 23.97 & 95.59 & 45.71 & - & - & - & - & - & - \\
    {GMMSeg}~\cite{liang2022gmmseg} & \xmark & 13.11  & 97.34  &43.47  &-  & - & - &47.90    &84.71   &34.42 \\
    {FED-U}~\cite{gudovskiy2023concurrent}& \xmark & 11.35  & 97.65  &37.05  &20.15  &95.96 &46.32 & -   & -  &- \\
    {FED-C}~\cite{gudovskiy2023concurrent}& \xmark & 18.48  & 96.34  &28.71  &32.69  &92.89 &25.34 & -   &-   &- \\
    \rowcolor{mygray}
    {SLEEG (ours)} & \xmark & \textbf{10.90} &\textbf{98.30}  & \textbf{70.90} & \textbf{3.85} & \textbf{99.22} & \textbf{77.23} & \textbf{42.60} & \textbf{86.60}  & \textbf{38.10}   \\
    \bottomrule
  \end{tabular}
  }
  \vspace{-.9em}
  \caption{Comparison on FS validation sets and Road Anomaly.
  ``OoD Data'' indicates the method adopts additional labeled data as anomalies for training. $\dagger$ indicates training with WideResNet38 backbone. \textbf{Bold} values represent the best results (Comparison is conducted within each group) .}
  \label{tab:fishyscapes_val}
  \vspace{-7mm}
\end{table*}

\section{Experiments}
\subsection{Datasets} 
We evaluate SLEEG in several widely used anomaly datasets: FishyScapes (FS) Lost \& Found~\cite{blum2021fishyscapes}, FishyScapes (FS) Static~\cite{blum2021fishyscapes}, Road Anomaly ~\cite{imgr}, Segment-Me-If-You-Can(SMIYC)\cite{2021SegmentMeIfYouCan} and StreetHazards~\cite{2019Scaling}.\\
\textbf{FS Lost \& Found} is a widely-used anomaly segmentation dataset with high-resolution samples for autonomous driving scenes. FS Lost \& Found is built based on the Lost \& Found~\cite{lostAndFound2016} and adopts the same collection setup as Cityscapes~\cite{cordts2016cityscapes}, which is also a segmentation benchmark for urban scenes. Specifically, 37 types of unexpected real road obstacles and 13 different street scenarios are included. 
A validation set of 100 samples is publicly available, while the hidden test set with 275 images is unknown. All the methods need to submit the code to the website\footnote{https://fishyscapes.com/results} to evaluate on this test set.  \\  
\textbf{FS Static} is artificially built upon the validation set of Cityscapes, where unexpected objects are collected from the Pascal VOC dataset~\cite{everingham2010pascal} and positioned randomly on the images. Specifically, only objects not belonging to the pre-defined classes of Cityscapes are used. This dataset consists of a public validation set with 30 images and a hidden test set with 1,000 images.\\  
\textbf{Road Anomaly} includes real-world images collected online, where anomalous obstacles encounter on or locate near the road. All the images are re-scaled to a size of 1,280 $\times$ 720 and pixel-wise annotations of unexpected objects are provided. 
Since there exists larger domain gap between Road anomaly and Cityscapes, generalization ability of models is essential to the performance on this dataset, following previous works~\cite{tian2021pixel,liang2022gmmseg}, we evaluate SLEEG on the test set consisting of $60$ images.  \\
\textbf{Other Benchmarks.} Besides the datasets above, we further evaluate on other benchmarks including SMIYC~\cite{2021SegmentMeIfYouCan} and Streethazards~\cite{2019Scaling}, which contain anomalies collected either from real-world or virtual game engine. The corresponding results can be found in supplementary material.

\vspace{-2mm}

\subsection{Implementation Details}
 For fair comparisons, we follow the similar settings of \cite{jung2021standardized,di2021pixel} to utilize the segmentation model of DeepLab series~\cite{chen2018encoder} with ResNet101~\cite{he2016deep} backbone, which is pre-trained on Cityscapes as in-distribution training and keep fixed without further re-training. The lightweight OoD head consists of three stacked Conv-BN-ReLU blocks and is trained for $40,000$ iterations with batchsize of $8$, the balance factor is set as $\lambda=0.5$ for test. Since the estimated anomalies score is not stable in the beginning and may affect the mask refinement, we apply a warmup strategy for mask refinement. First we take JEM as anomaly score for $6,000$ iterations and then replace it with estimated anomaly likelihood. Following common setting \cite{jung2021standardized, blum2021fishyscapes, xia2020synthesize}, the average precision (AP), area under receiver operating characteristics curve (AUROC) and false positive rate ($\rm{FPR_{95}}$) at true positive rate of 95\%  are adopted as metrics to perform comprehensive evaluation. Among these metrics, $\rm{FPR_{95}}$ and AP are more crucial since there exists severe imbalance between anomaly and normal pixels.

\subsection{Experiments Results}

\subsubsection{Comparison with State-of-the-art Methods}

\textbf{FS Leaderboard.} Tab.~\ref{tab:FS_leaderboard} provides results on test sets of FS benchmark. Following ~\cite{blum2019fishyscapes}, previous methods are categorized based on whether they require re-training the segmentation network or extra OoD data. Compared with previous methods without re-training or extra anomaly data, SLEEG works effectively to outperform most competitors by large margins. 
SLEEG can even surpass some methods with re-training and extra anomaly data~\cite{tian2021pixel,synboost2021} by large margins and achieve a new SOTA performance of AP on Lost \& Found track, which includes anomalies in more realistic scenes. On the artificially-generated FS Static, SLEEG is on par with the SOTA method DenseHybrid, which requires extra data and adopts stronger Wide-ResNet as its feature extractor. Although SLEEG is inferior to some methods relying on extra data in FS Static, this is due to the external data~\cite{lin2014microsoft} adopted falls into similar distribution of anomalies in FS Static~\cite{everingham2010pascal}, thus they suffer a large performance gap between artificial Static track and realistic LAF, while SLEEG keeps superiority on both tracks.

\noindent\textbf{FS Validation Sets.} Tab.~\ref{tab:fishyscapes_val} shows comparisons on the validation sets of FS Lost \& Found and Static. In terms of AP, our anomaly detector achieves the best results for all metrics on FS Lost \& Found and competitive performance on FS Static. Specifically, when compared with GMMSeg~\cite{liang2022gmmseg}, SLEEG yields significant improvements of 26.4\% on AP and reduce $\rm{FPR_{95}}$ to 10.9\%. The results demonstrate the effectiveness and robustness of our SLEEG on detecting real-world anomaly instances. Similar to results in  Tab.~\ref{tab:FS_leaderboard}, methods with auxiliary anomaly data (\emph{e.g.} PEBAL) or synthetic model (\emph{e.g.} SynthCP) usually suffer from significant performance gap between artificial anomaly in FS Static and anomalies in realistic scenes from LAF, since they tend to overfit label anomalies. In contrast, SLEEG can yield relative consistent performance gain.  

\noindent\textbf{Road Anomaly Test Sets.} We further compare SLEEG with recent advanced anomaly segmentation methods on Road Anomaly in Tab.~\ref{tab:fishyscapes_val}, it is observed that SLEEG outperforms most competitors by a large margin when no labeled anomaly data is available. Since there exists larger inherent domain shift between Road anomaly and Cityscapes than Fishyscapes, previous methods (\emph{e.g.} SML~\cite{jung2021standardized}) that perform well on Fishyscapes are prone to suffer from poor accuracy on Road anomaly. However, our SLEEG yields significant improvements on both two datasets, demonstrating the robustness of SLEEG in tackling open-world scenes with diverse styles. 

\begin{table}
  \centering
  \scriptsize
  \setlength{\tabcolsep}{0.7mm}{
  \begin{tabular}{cc|ccc|ccc}
    \toprule
    \multicolumn{2}{c|}{Estimator} & \multicolumn{3}{c}{FS LAF Validation} &  \multicolumn{3}{|c}{Road Anomaly Test}\\
    \cline{1-8} TAE & TORE & $\rm{FPR_{95}}$ $\downarrow$ & AUROC $\uparrow$ & AP $\uparrow$ & $\rm{FPR_{95}}$ $\downarrow$ & AUROC $\uparrow$ & AP $\uparrow$ \\
    \midrule
    \midrule
    {\XSolidBrush} & {\XSolidBrush} & 27.0  &95.5    &31.0  &79.9    &71.3   &18.9 \\
    {\Checkmark} & {\XSolidBrush} &  25.7 &96.2   &57.5  & 47.6   & 82.9  & 33.5  \\
    {\XSolidBrush}& {\Checkmark} & 18.4  & 96.9  & 47.9  & 51.2 & 82.8 & 31.6    \\
     {\Checkmark}& {\Checkmark} & \textbf{10.9}  & \textbf{98.3}  & \textbf{70.9}  &\textbf{42.6} & \textbf{86.6} & \textbf{38.1}   \\
    \bottomrule
  \end{tabular}
  }
  \vspace{-1em}
  \caption{
  Ablation results for different likelihood estimators on FS validation set and Road Anomaly validation set. }
  \label{tab:ablation_component}
  \vspace{-4mm}
\end{table}



\section{In-depth Discussion}
\label{sec:ablation}

 \begin{table}
  \centering
  \scriptsize
  \setlength{\tabcolsep}{0.7mm}{
  \begin{tabular}{c|c|ccc|ccc}
    \toprule
    \multirow{2}*{\makecell[c]{Patch\\Policy}} & \multirow{2}{*}{\makecell[c]{Refine \\ ment}} & \multicolumn{3}{c}{FS LAF Validation} &  \multicolumn{3}{|c}{Road Anomaly Test}\\
    \cline{3-8} & & $\rm{FPR_{95}}$ $\downarrow$ & AUROC $\uparrow$ & AP $\uparrow$ & $\rm{FPR_{95}}$ $\downarrow$ & AUROC $\uparrow$ & AP $\uparrow$ \\
    \midrule
    \midrule
    \makecell[c]{Void \\as OoD} & - & 19.3   & 96.8    & 47.3  &  69.1   &72.5    & 21.3 \\
    \hline
    \multirow{2}{*}{\makecell[c]{Square\\Patch}} & \xmark & 25.4   & 95.1    & 45.7  & 67.9    & 75.0    & 18.2 \\
    & \cmark & \textbf{13.8}   & \textbf{97.5}    & \textbf{62.5}  & \textbf{47.8}    &\textbf{83.9}    & \textbf{34.8} \\
    \hline
    \multirow{2}{*}{\makecell[c]{Convex\\Patch}} & \xmark & 13.3   & 97.9    & 61.9  & 44.8    & 84.6    & 35.9 \\
     & \cmark & \textbf{10.9}  & \textbf{98.3}  & \textbf{70.9}  & \textbf{42.6} & \textbf{86.6} & \textbf{38.1} \\
    \bottomrule
  \end{tabular}
  }
  \vspace{-1em}
  \caption{
  Investigation of patch policies on FS and Road Anomaly validation set. ``Void as OoD" denotes training with pixels of void class in Cityscapes as anomaly samples.}
  \vspace{-4mm}
  \label{tab:ablation_shape}
\end{table}

\noindent\textbf{Effectiveness of Different Estimators.} We explore the contribution of TAE and TORE on the anomaly performance of FS Lost \& Found and Road Anomaly test sets. To make comparison, we take the pure JEM~\cite{grathwohl2019your} measurement as baseline anomaly detector. It can be observed from Tab.~\ref{tab:ablation_component} that TAE brings significant improvements to the baseline on the AP by training with self-generated OoD samples. Further, by dynamically adjusting the margin of anomaly scores, TORE also brings a large performance boost of around 13\% in AP and significant reduction in $\rm{FPR_{95}}$ as well. Finally, SLEEG is capable of achieving the best performance by seamlessly integrating two estimators for the prediction of anomaly areas. 

Specifically, as indicated by Eq.~(\ref{eq:dynamic}), we also compare the automatic energy-guided margin $\widehat{\gamma}$ with a manually tuned static value $\gamma$ for the training of TORE, which is equivalent to eliminating the JEM terms in Eq.~(\ref{eq:dynamic}). For both setting, we set $\gamma=15$, the results can be illustrated in Tab.~\ref{tab:ablation_gamma}. It is clear that the margin with a dynamically controlled component as Eq.~(\ref{eq:dynamic}) can consistently outperforms static margin, indicating the residual form of anomaly likelihood can adaptively capture effective training samples.

\begin{figure}[!t]
 \centering
\scriptsize
\hspace{-5mm}
 \begin{minipage}{0.49\textwidth}
    \centering
    \includegraphics[width=0.95\textwidth]{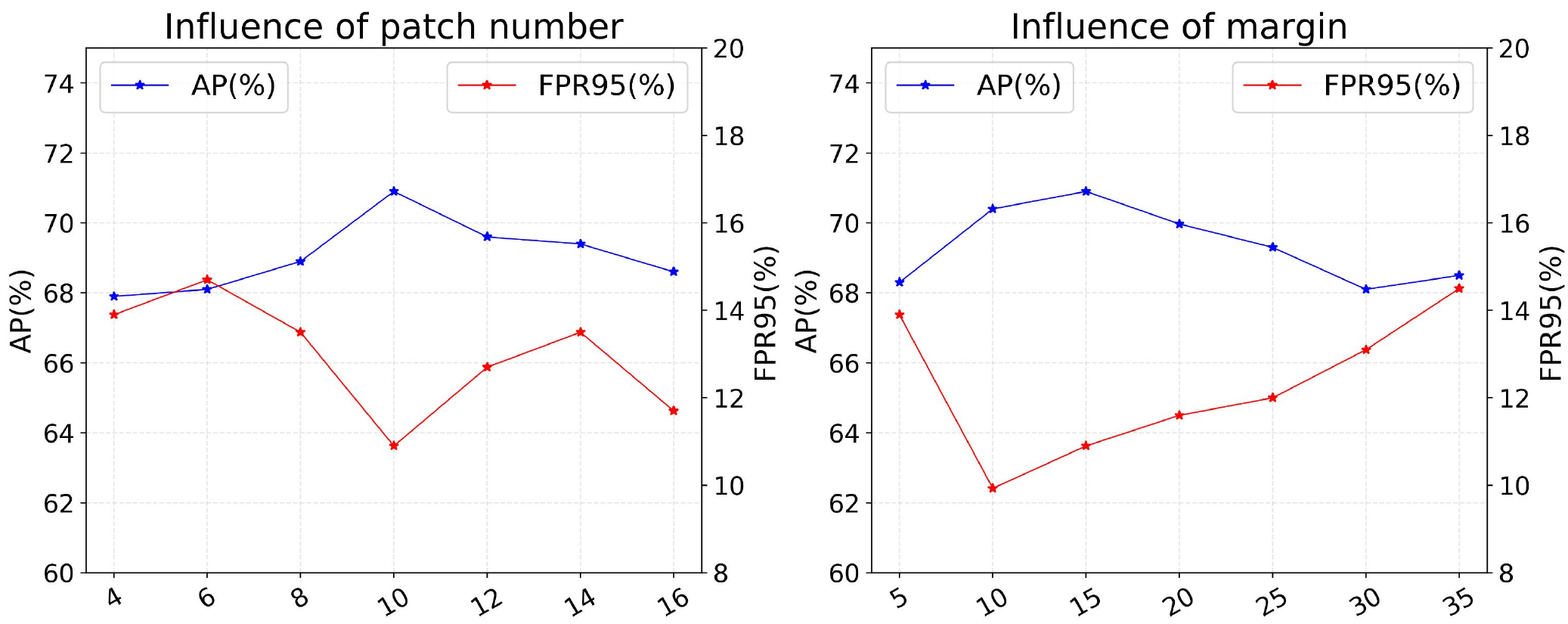}
\end{minipage}
    \vspace{-3mm}
  \caption
    {
        Investigation on the influence on AP and false positive rate with varied patch number $N$ (left) and margin value $\gamma$ (right) on FS Lost \& Found validation set. 
      }
  \label{fig:patchnum}
  \vspace{-7mm}
 \end{figure}

 \begin{table}
  \centering
  \scriptsize
  \setlength{\tabcolsep}{0.9mm}{
  \begin{tabular}{c|c|cc|cc|c}
    \toprule
    \multirow{2}*{Architecture} &\multirow{2}*{Method} & \multicolumn{2}{c|}{FS LAF Val} & \multicolumn{2}{|c|}{FS Static Val}  & \multirow{2}*{mIoU}\\
    \cline{3-6} &  & $\rm{FPR_{95}}$~$\downarrow$& AP~$\uparrow$   & $\rm{FPR_{95}}$~$\downarrow$ & AP~$\uparrow$ \\
    \midrule
    \midrule
    \multirow{4}{*}{\makecell[c]{OCRNet\\~\cite{YuanCW20}}}& SML  & 18.28  & 39.96 &  15.07 &  47.90 & \multirow{4}*{80.66} \\
    & JEM  &  22.90  & 23.27 &  16.80 & 34.03 & \\
    & Void Class &  16.65  &  46.62  & 17.74  & 29.30 & \\
    & SLEEG  & \textbf{8.89} & \textbf{72.51} & \textbf{7.6} & \textbf{73.01} &  \\ \hline
    \multirow{4}{*}{\makecell[c]{ISANet\\~\cite{huang2019isa}}}& SML  & 18.67  & 28.76 &  14.86 &  32.15 & \multirow{4}*{80.51} \\
    & JEM   &  35.57 & 22.65 &  16.22 & 45.22& \\
    & Void Class &  29.17 & 43.53 &  17.18 & 24.61& \\
    &SLEEG  & \textbf{12.28} & \textbf{65.79} & \textbf{3.33} & \textbf{80.03} & \\ \hline
    \multirow{4}{*}{\makecell[c]{FCN\\~\cite{shelhamer2017fully}}}   & SML  &  39.80  &  17.59 &  14.53  &  28.92 & \multirow{4}*{77.72} \\
    & JEM   & 39.36 &  21.83  &  13.47 & 30.51 & \\
    & Void Class &  24.29  &    43.44 &  14.79 & 22.72& \\
    &SLEEG & \textbf{21.01} &  \textbf{63.11} &  \textbf{5.31}  & \textbf{61.29} &\\ \hline

    \bottomrule
  \end{tabular}
  }
  \vspace{-3mm}
  \caption{
  Comparison between SLEEG and other anomaly detection methods on FS validation sets with different segmentation models. "Void Class" denotes training models with pixels that fall into void class as anomaly samples.}
    \vspace{-2mm}
  \label{tab:fishyscapes_framework}
\end{table}
\begin{table}
\vspace{-2mm}
  \centering
  \scriptsize
  \setlength{\tabcolsep}{0.7mm}{
  \begin{tabular}{c|ccc|ccc}
    \toprule
    \multirow{2}{*}{Margin Type} & \multicolumn{3}{c}{FS LAF Validation} &  \multicolumn{3}{|c}{Road Anomaly Test}\\
    \cline{2-7} & $\rm{FPR_{95}}$ $\downarrow$ & AUROC $\uparrow$ & AP $\uparrow$ & $\rm{FPR_{95}}$ $\downarrow$ & AUROC $\uparrow$ & AP $\uparrow$ \\
    \midrule
    \midrule
     {Static $\gamma$} & 12.3  & 97.1 & 65.2  & 56.2 & 81.6  & 34.4 \\
     {Dynamic $\widehat{\gamma}$} & \textbf{10.9} & \textbf{98.3}  & \textbf{70.9}  & \textbf{42.6} & \textbf{86.6} & \textbf{38.1}  \\
    \bottomrule
  \end{tabular}
  }
  \vspace{-1em}
  \caption{
  Ablation results of comparing static/dynamic margin (Eq.~(\ref{eq:dynamic})) on FS and Road Anomaly validation set. }
  \vspace{-6mm}
  \label{tab:ablation_gamma}
\end{table}

\noindent\textbf{Investigation on Patch Policy.} Tab.~\ref{tab:ablation_shape} shows the comparison results for different patch policies on FS Lost \& Found validation set and road anomaly validation set. Especially, we design a baseline which regards pixels labeled with ``void'' from Cityscapes as auxiliary anomaly data (denoted as ``Void as OoD''). From Tab.~\ref{tab:ablation_shape}, training with simple square patches performs slightly inferior to baseline, since this is prone to implicitly learn useless shape-related prior. With more various shapes, convex patches brings significant improvements. Besides, by performing our refinement strategy, SLEEG can also achieve significant improvement even with simple square patch. Finally, imposing further refinement on the convex patch achieves the best performance.     

\noindent\textbf{Parameter Sensitivity.} We further investigate the influence of patch number and margin value on the FS Lost \& Found validation set. As shown in Fig.~\ref{fig:patchnum}, SLEEG performs best when margin $\gamma=15$ and patch number $N=10$ . The results demonstrate that SLEEG achieves similar performance across all settings, implying SLEEG is not very sensitive to both patch number and margin value. Finally, we set patch number and margin to 10 and 15 respectively.

\noindent\textbf{Extension to More Segmentation Models.}
Finally, as shown in Tab.~\ref{tab:fishyscapes_framework}, we also validate the generalization ability of SLEEG by training with different advanced segmentation frameworks, including OCRNet, ISANet and FCN, where SLEEG consistently surpasses SML and JEM across all the frameworks with more than $20\%$ AP improvement on average. Additionally, it is observed that SLEEG performs better on FS Lost\& Found validation set when adopting networks with higher mIoU scores, implying that SLEEG can work effectively with various frameworks. More detailed results can be found in the supplementary material.

\section{Conclusion}
In this paper, we propose SLEEG, a simple and flexible anomaly segmentation model without re-training or labeled anomaly data, which exploits a task-agnostic binary estimator, and a task-oriented energy residual estimator for
anomaly likelihood estimation, and incorporate them with an adaptive copy-and-paste mask policy for self-supervised learning. Extensive experimental results verify the effectiveness of our method and competitive performance is achieved on both FS Lost \& Found validation and test sets by SLEEG. 


\section*{Acknowledgement}
This work was supported by National Natural Science Fund of China (62076184, 61976158, 61976160, 62076182, 62276190), in part by Fundamental Research Funds for the Central Universities and State Key Laboratory of Integrated Services Networks (Xidian University), in part by Shanghai Innovation Action Project of Science and Technology (20511100700) and Shanghai Natural Science Foundation (22ZR1466700).

\section*{Appendix}

\section{More Experimental Results}

 \begin{table*}
  \centering
  \setlength{\tabcolsep}{0.9mm}{
  \begin{tabular}{l|c|cc|cc|c}
    \toprule
    
    \multirow{2}*{Architecture} &\multirow{2}*{Method} & \multicolumn{2}{c|}{FS LAF Val} & \multicolumn{2}{|c|}{FS Static Val}  & \multirow{2}*{mIoU}\\
    \cline{3-6} &  & $\rm{FPR_{95}}$~$\downarrow$& AP~$\uparrow$   & $\rm{FPR_{95}}$~$\downarrow$ & AP~$\uparrow$ \\
    \midrule
    \midrule
    \multirow{4}{*}{OCRNet~\cite{YuanCW20}}& SML  & 18.28  & 39.96 &  15.07 &  47.90 & \multirow{4}*{80.66} \\
                                                    & JEM  &  22.90  & 23.27 &  16.80 & 34.03 & \\
                                                    & Void Class &  16.65  &  46.62  & 17.74  & 29.30 & \\
                                                    & SLEEG  & \textbf{8.89} & \textbf{72.51} & \textbf{7.6} & \textbf{73.01} &  \\ \hline
    \multirow{4}{*}{ISANet~\cite{huang2019isa}}& SML  & 18.67  & 28.76 &  14.86 &  32.15 & \multirow{4}*{80.51} \\
                                                    & JEM   &  35.57 & 22.65 &  16.22 & 45.22& \\
                                                    & Void Class &  29.17 & 43.53 &  17.18 & 24.61& \\
                                                    &SLEEG  & \textbf{12.28} & \textbf{65.79} & \textbf{3.33} & \textbf{80.03} & \\ \hline
    \multirow{4}{*}{\makecell{FCN~\cite{shelhamer2017fully}}}   & SML  &  39.80  &  17.59 &  14.53  &  28.92 & \multirow{4}*{77.72} \\
                                                    & JEM   & 39.36 &  21.83  &  13.47 & 30.51 & \\
                                                    & Void Class &  24.29  &    43.44 &  14.79 & 22.72& \\
                                                    &SLEEG & \textbf{21.01} &  \textbf{63.11} &  \textbf{5.31}  & \textbf{61.29} &\\ \hline

    \bottomrule
  \end{tabular}
  }
  \caption{Comparison between SLEEG and other anomaly detection methods on FS validation sets with different segmentation models. "Void Class" denotes training models with pixels that fall into void class as anomaly samples.}
  \label{tab:fishyscapes_framework}
\end{table*}

\begin{figure*}[!t]
 \centering
\small
 \begin{minipage}{0.97\textwidth}
    \centering
    \includegraphics[width=\textwidth]{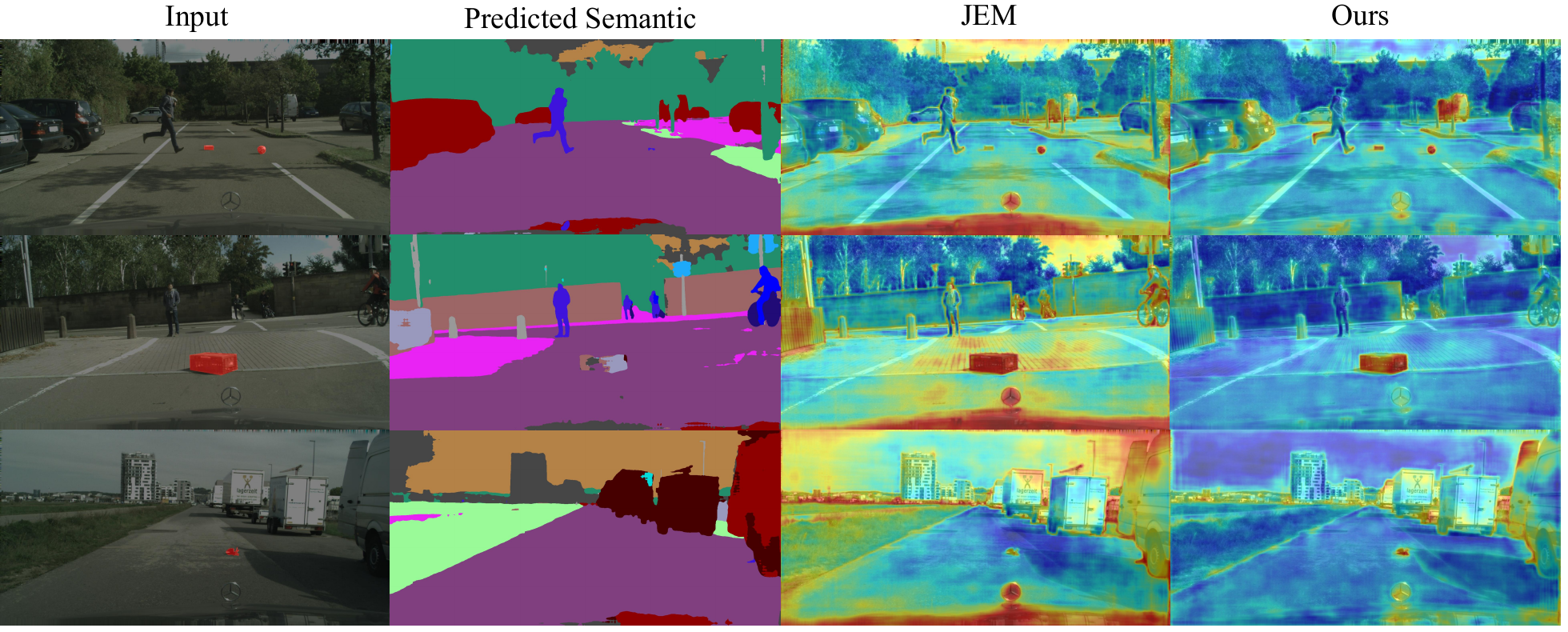}
\end{minipage}
\vspace{-1em}
  \caption
    {
    \small
        Visualization of  on FS Lost \& Found validation set. Compared with JEM, predictions from our SLEEG show anomaly maps with higher responses for anomalous instances and lower values for normal pixels.
      }
  \label{fig:visualize}
 \end{figure*}

\textbf{Results on Lost \& Found.} To make more comprehensive comparison, we also evaluate our method on the old Lost \& Found dataset~\cite{lostAndFound2016}, where 13 challenging real-world scenes are included with 37 different anomalous instances, which is the first publicly available urban anomaly segmentation datasets. Specifically, the obstacles in this dataset vary significantly in size and material. And following the official protocol~\cite{lostAndFound2016}, 1,203 images collected from 112 video stereo sequences with a resolution of 2048$\times$1024 are utilized as the test set. Similar to the setting on FS Lost \& Found, the segmentation model of DeepLab series~\cite{chen2018encoder} with ResNet101 pretrained on Cityscapes is used and fixed without re-training on this dataset. Finally, the same OoD head and evaluation metrics are adopted as FS Lost \& Found.

\begin{table*}[t!]
\centering
\resizebox{1.0\linewidth}{!}{%
\begin{tabular}{@{}c|cc|ccc@{}}
\toprule
Methods & \shortstack{OoD Data}&\shortstack{Re-training}     & AUC $\uparrow$ & AP $\uparrow$ & FPR$_{95}$ $\downarrow$ \\ \hline
  \midrule
Meta-OoD~\cite{chan2021entropy} & \cmark & \cmark & 97.95  &  71.23   & 5.95   \\

SynBoost$^{\dagger}$~\cite{di2021pixel} & \cmark &\xmark  & 98.38 & 70.43  & 4.89    \\
Deep Gambler$^{\dagger}$~\cite{liu2019deep} &\cmark  &\cmark  & 98.67   & 72.73  & 3.81    \\
PEBAL$^{\dagger}$~\cite{di2021pixel}  &\cmark &\cmark  & \textbf{99.76} & \underline{78.29} & \textbf{0.81}  \\ 
DenseHybrid$^{\dagger}$~\cite{grcic2022densehybrid} & \cmark & \cmark & \underline{99.37} & \textbf{78.70} & \underline{2.10}  \\ \midrule

MSP~\cite{hendrycks2019scaling}  & \xmark & \xmark & 85.49 & 38.20 & 18.56  \\
Mahalanobis~\cite{lee2018simple} & \xmark & \xmark & 79.53  & 42.56  & 24.51  \\
Max Logit~\cite{hendrycks2016baseline} &\xmark & \xmark& \underline{94.52}  & 65.45  & \underline{15.56}  \\
Entropy~\cite{hendrycks2016baseline} &\xmark  & \xmark & 86.52  & 50.66  & 16.95  \\
SML$^{\dagger}$~\cite{jung2021standardized}&\xmark  &\xmark  & 88.05 & 25.89  & 44.48    \\
Energy~\cite{liu2020energy}  &\xmark  &\xmark & 94.45  & \underline{66.37}  & 15.69    \\
\rowcolor{mygray}
SLEEG & \xmark & \xmark  & \textbf{98.59} & \textbf{82.88} & \textbf{4.60} \\

\bottomrule
\end{tabular}%
}
\caption{Comparison on \textbf{Lost \& Found} testing set. All methods use the same segmentation models. $\dagger$ indicates training with WideResNet38 backbone.  \textbf{Bold} values and \underline{underlined} values represent the best and second best results.
}
\label{tab:lostfoundtrue}

\end{table*}

 Tab.~\ref{tab:lostfoundtrue} shows the performance of the proposed SLEEG on the test set of Lost \& Found. Notably, our method surpasses all previous approaches and achieves SOTA performance on AP. Specifically, though previous SOTA methods DenseHybrid~\cite{grcic2022densehybrid} and PEBAL~\cite{tian2021pixel} utilize auxiliary data and are re-trained with more complex WideResNet38 as backbone, our SLEEG still brings a relative improvement of 4.59\% and 4.18\% to them on AP respectively with ResNet101 as backbone and requires no auxiliary OoD samples or further re-training of the segmentation model. Furthermore, when compared with the SOTA method Energy~\cite{liu2020energy} that falls into the same category as SLEEG, our method surpasses it by a large gap of 16.51\% on AP and meanwhile significantly reduces the false positive pixels from 15.69\% to 4.6\%. This results fully illustrate that SLEEG shows great robustness and effectiveness on localising various real-world unexpected objects. Moreover, SML~\cite{jung2021standardized} attempts to tackle this task by re-balance the class-wise discrepancy of inlier samples. However, since there are only two classes in Lost \& Found, there exists large performance gap between the performance of SML on Fishyscapes and Lost \& Found, implying that SML may not be suitable for real-world applications. By contrast, it is worth noting that our SLEEG is capable of achieving consistent performance boost on all these datasets, demonstrating its great potential in real world scenarios. 

 \begin{table}
  \centering
  \footnotesize
  \setlength{\tabcolsep}{0.7mm}{
  \begin{tabular}{c|ccc}
    \toprule
    \multirow{2}*{Method} & \multicolumn{3}{c}{FS LAF Validation} \\
    \cline{2-4} & $\rm{FPR_{95}}$ $\downarrow$ & AP $\uparrow$ & AUROC $\uparrow$\\
    \midrule
    \midrule
    \makecell{DenseHybrid + OoD data\\ ~\cite{grcic2022densehybrid}} & 6.1 & 63.8 & -  \\
    \hline
    DenseHybrid + Convex patch  & 22.2  & 56.2 & 96.6 \\
    DenseHybrid + Refined convex patch & 18.6 & 60.5 & 97.0 \\ \hline
    SLEEG + Refined convex patch & 10.9 & 70.9 & 98.3 \\
    \bottomrule
  \end{tabular}
  }
  \caption{Investigation when combining other state-of-the-art method DenseHybrid~\cite{grcic2022densehybrid} with our strategy of patch generation on FS LAF validation set.}
  \label{tab:ablation_shape2}
\end{table}

\begin{table*}[h]
\begin{center}
\caption{Performance evaluation on the SMIYC benchmark \cite{chan21arxiv}.
}
\label{table:smiyc}
\begin{tabular}{l|c|c|cc}
\hline 
\multicolumn{1}{l|}{\multirow{2}{*}{Method}} & \multicolumn{1}{c|}{\multirow{1}{*}{OoD}} & \multicolumn{1}{c|}{\multirow{1}{*}{Re-}} & \multicolumn{2}{c}{AnomalyTrack} \\ 
\multicolumn{1}{l|}{} & \multicolumn{1}{c|}{Data} & \multicolumn{1}{c|}{training}  & \multicolumn{1}{c}{$\mathrm{FPR}_{95}$} $\downarrow$ & AP $\uparrow$\\\hline \hline
SynBoost \cite{biase21cvpr} & \cmark& \xmark  & 61.9 & \underline{56.4}\\
JSRNet \cite{vojir21iccv} & \xmark&  \cmark & 43.9 & 33.6 \\
Void Classifier \cite{blum21ijcv} & \cmark& \cmark  & 63.5 & 36.6\\
DenseHybrid~\cite{grcic2022densehybrid} & \cmark&  \cmark  & \textbf{9.8}& \textbf{78.0} \\
PEBAL~\cite{tian2021pixel} &\cmark  &\cmark  & \underline{40.8} & 49.1 \\ \hline

Image Resyn. \cite{lis19iccv} & \xmark& \xmark& \textbf{25.9}& \underline{52.3} \\
Embed.\ Dens.\ \cite{blum21ijcv} & \xmark&  \xmark  & 70.8 & 37.5 \\
ODIN \cite{liang18iclr} & \xmark&  \xmark  & 71.7 & 33.1\\
MC Dropout \cite{kendall17nips} & \xmark& \xmark  & 69.5 & 28.9\\
Max softmax \cite{hendrycks17iclr} & \xmark&  \xmark & 72.1 & 28.0  \\
Mahalanobis \cite{lee18nips} & \xmark&  \xmark  & 87.0 & 20.0 \\
\rowcolor{mygray}
SLEEG (ours) & \xmark&  \xmark  & \underline{36.4}  & \textbf{52.7}\\ \hline
\end{tabular}
\end{center}
\end{table*}

\begin{table*}[h]
\begin{center}
\caption{Performance evaluation on StreetHazards \cite{streethazards2019}.
SLEEG achieves competitive anomaly detection performance.}
\label{table:osr_sh}
\begin{tabular}{l|c|c|ccc}
\hline 
\multirow{2}{*}{Method} & \multicolumn{1}{c|}{OoD} & \multicolumn{1}{c|}{Re-} & \multicolumn{3}{c}{Anomaly detection}\\  \cline{4-6} 
  & \multicolumn{1}{c|}{Data} & \multicolumn{1}{c|}{training}        & $\mathrm{FPR}_{95}$ $\downarrow$       & \multicolumn{1}{c}{AUROC $\uparrow$ }  &  AP $\uparrow$ \\ \hline \hline

Energy \cite{liu20neurips}& \cmark & \cmark & 18.2  & 93.0 &  12.9 \\
Outlier Exposure \cite{hendrycks19iclr} & \cmark & \cmark    &  {17.7}  & {94.0} &  14.6\\
OOD-Head \cite{bevandic19gcpr}  & \cmark & \cmark  &  56.2   & 88.8 &  {19.7}\\
OH-MSP \cite{bevandic21arxiv} & \cmark  &\cmark   & 30.9 & 89.7& 18.8 \\
DenseHybrid~\cite{grcic2022densehybrid} & \cmark &\cmark   & \underline{13.0} & \underline{95.6} &  \underline{30.2}  \\ 
CoroCL~\cite{liu2023residual} & \cmark &\cmark   & \textbf{8.2} & \textbf{97.2} &  \textbf{31.2}  \\

\hline

SynthCP \cite{xia20eccv} & \xmark & \xmark           & 28.4          & 88.5 &  9.3 \\
Dropout \cite{kendall17nips}\cite{xia20eccv} &  \xmark &\xmark           & 79.4          & 69.9&  7.5 \\
TRADI \cite{franchi20eccv} & \xmark &\xmark           & 25.3          & 89.2&  7.2   \\
OVNNI \cite{franchi20arxiv} & \xmark &\xmark  & 22.2 & 91.2  & 12.6\\
SO+H \cite{grcic21visapp}& \xmark &\xmark  & 25.2 & 91.7 & 12.7  \\
DML \cite{cen21iccv} & \xmark & \xmark  & \underline{17.3}  &  \underline{93.7} & \underline{14.7} \\
MSP \cite{hendrycks17iclr} & \xmark &\xmark    &   27.9  & 90.1&  7.5 \\
ML \cite{hendrycks19arxiv} & \xmark &\xmark    &  22.5    & 92.4& 11.6 \\
ODIN \cite{liang18iclr}&   \xmark & \xmark        & 28.7    &   90.0 &    7.0  \\
ReAct \cite{sun21neurips} & \xmark &\xmark  & 21.2 & 92.3 & 10.9 \\
\rowcolor{mygray}
SLEEG (ours) & \xmark &\xmark   & \textbf{13.3} & \textbf{95.7}  &  \textbf{27.6} \\\hline
\end{tabular}
\end{center}
\end{table*}

\begin{figure}[!t]
 \centering
\small
 \begin{minipage}{0.5\textwidth}
    \centering
    \includegraphics[width=\textwidth]{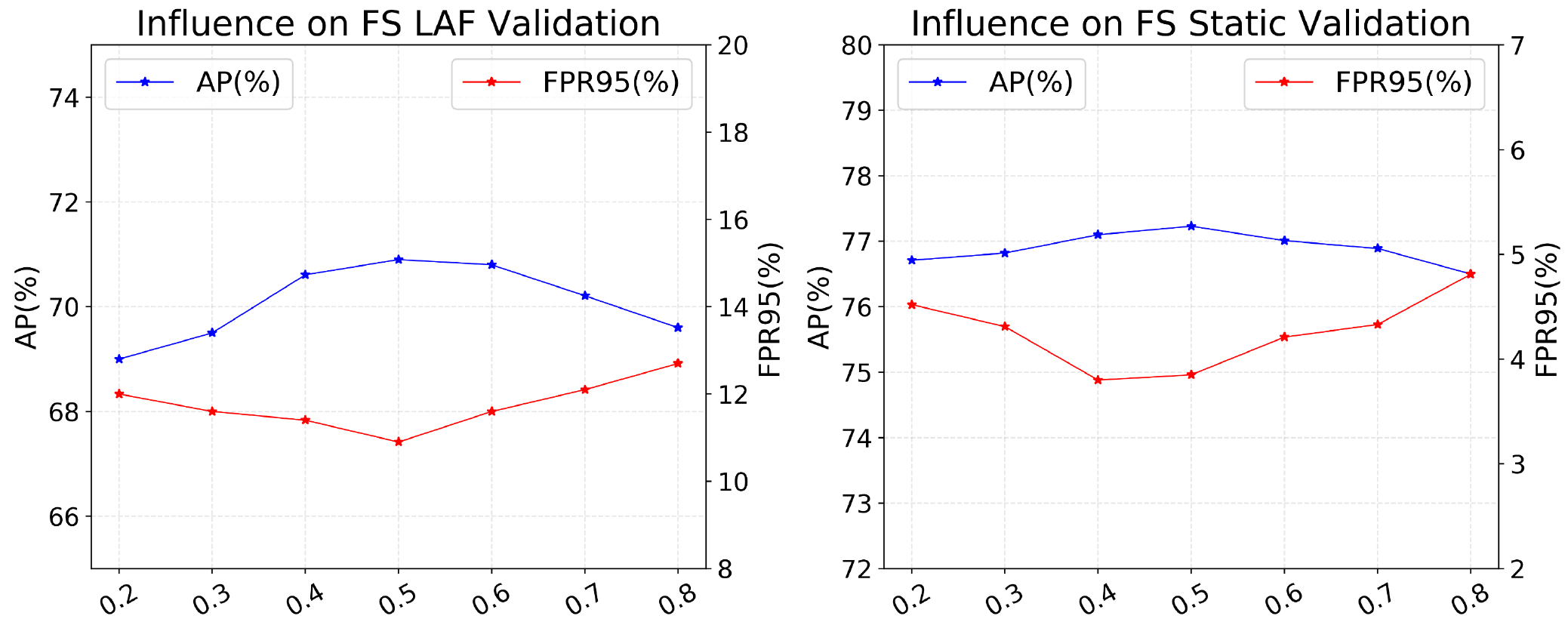}
\end{minipage}
\vspace{-1em}
  \caption
    {
    \small
        Investigation on the influence on AP and false positive rate with varied $\lambda$ value on FS Lost \& Found validation set (left) and FS Static validation set (right). 
      }
  \label{fig:labmda}

 \end{figure}

\textbf{Results for OoD data replacement.} Meanwhile, we also separately investigate our patch policy with other method~\cite{grcic2022densehybrid} by replacing the external OoD data used in original paper with our generated patches. The results can be illustrated in Tab.~\ref{tab:ablation_shape2}, where the results of Densehybrid+OoD data is borrowed from the original paper, and the other results are reimplemented by ourselves. It can be observed that although applying less data, our patch policy can achieve results close to those in the original paper, this further demonstrate that our scheme can be viewed as a potential replacement for OoD data relied by previous methods.

\textbf{Parameter Sensitivity.} Finally we investigate the influence of the balance factor $\lambda$ on the FS Lost \& Found validation set and FS Static validation set. As shown in Fig.~\ref{fig:labmda}, SLEEG performs best when the balance factor $\lambda=0.5$. The results demonstrate that SLEEG achieves similar performance across all settings, implying SLEEG is not very sensitive to $\lambda$. Finally, we set $\lambda=0.5$.

\begin{figure}[!t]
 \centering
\small
 \begin{minipage}{0.49\textwidth}
    \centering
    \includegraphics[width=0.95\textwidth]{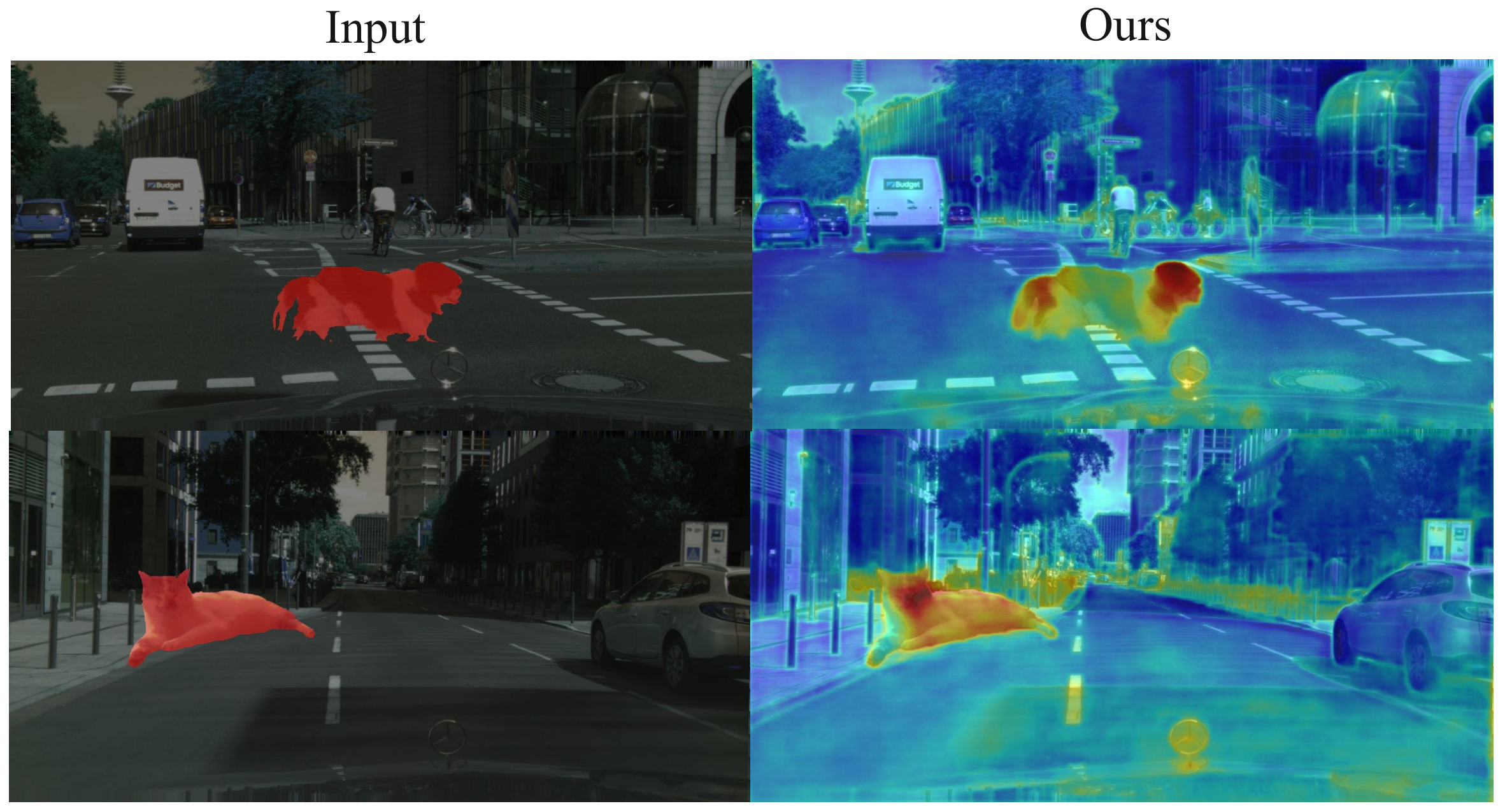}
\end{minipage}

  \caption
    {
    \small
        Visualization results of some failure cases from SLEEG on FS Static validation set.  
      }
  \label{fig:densehybrid_static}

 \end{figure}

\section{Qualitative Results}
\begin{figure*}[!t]
 \centering
\small
 \begin{minipage}{0.99\textwidth}
    \centering
    \includegraphics[width=\textwidth]{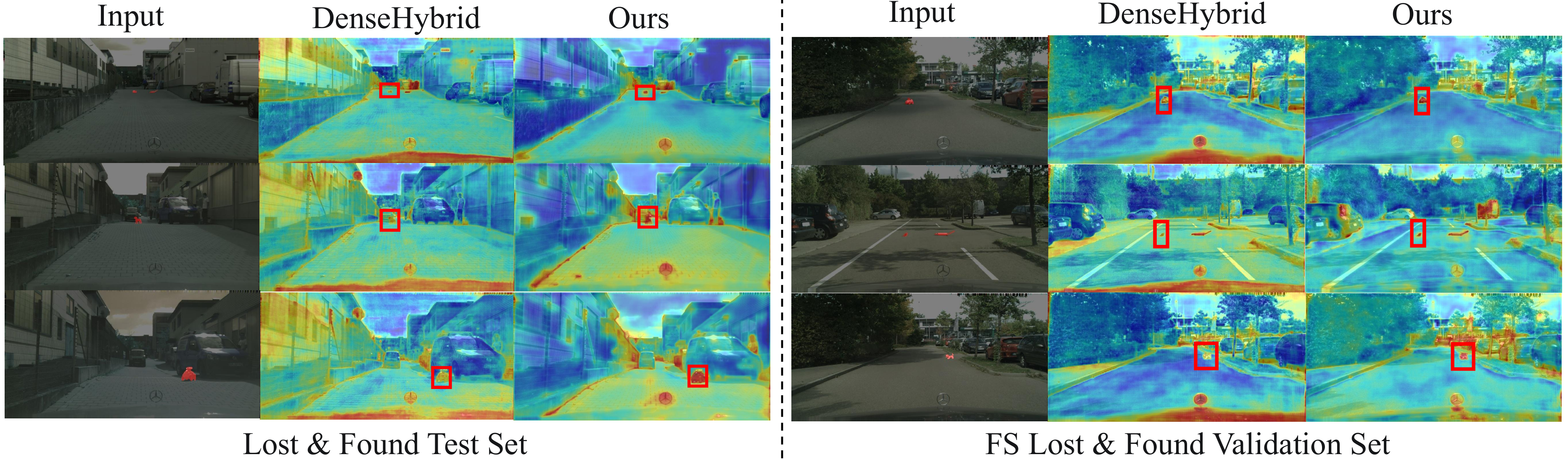}
\end{minipage}

\vspace{-1em}
  \caption
    {
    \small
        Comparison of visualization results with Densehybrid on Lost \& Found test set and FS Lost \& Found validation set. The proposed SLEEG is capable of localizing small anomalous objects more accurately. For example, SLEEG generates higher anomaly scores for the instances in the red box of the second and third rows. 
      }
  \label{fig:densehybrid_visualize_fishy}

 \end{figure*}

We also conduct further comparison with the SOTA method DenseHybrid on Lost \& Found test set and FS Lost \& Found validation set. As shown in Fig.~\ref{fig:densehybrid_visualize_fishy}, for the challenging small anomalous objects at a distance, SLEEG produces larger anomaly scores, allowing vehicles respond to prevent accidents in a more timely manner. Besides, for objects with diverse shapes, our SLEEG is capable of producing a more complete contour covering the entire anomalous object as shown in the third row of Fig.~\ref{fig:densehybrid_visualize_fishy}. 

\begin{figure*}[!t]
 \centering
\small
 \begin{minipage}{0.99\textwidth}
    \centering
    \includegraphics[width=\textwidth]{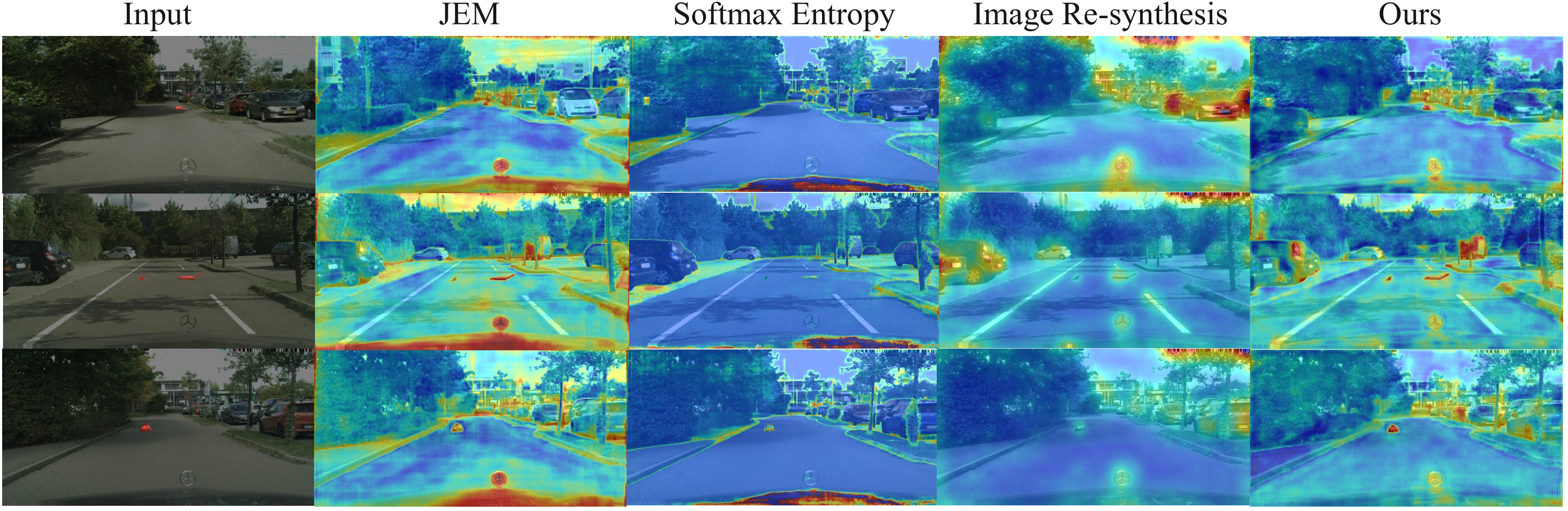}
\end{minipage}

  \caption
    {
    \small
        Comparison of visualization results with JEM, Softmax Entropy and Image Re-synthesis on FS Lost \& Found validation set.  
      }
  \label{fig:resynthesis_fslaf}

 \end{figure*}

\begin{figure*}[!t]
 \centering
\small
 \begin{minipage}{0.99\textwidth}
    \centering
    \includegraphics[width=\textwidth]{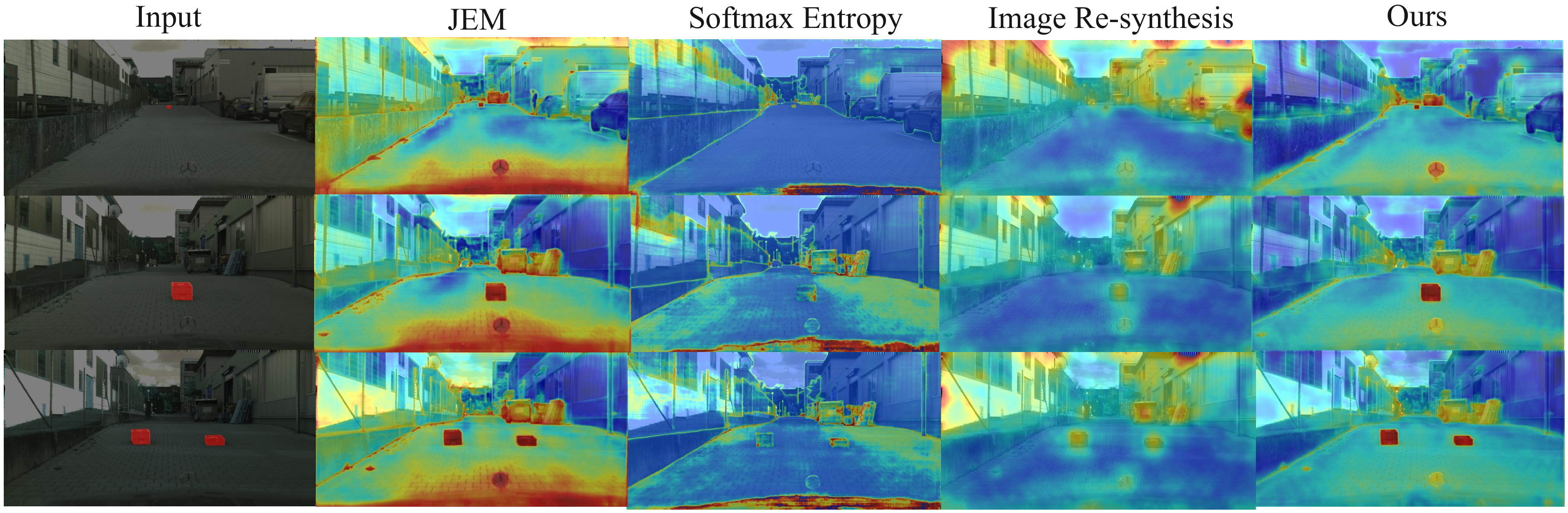}
\end{minipage}

  \caption
    {
    \small
        Comparison of visualization results with JEM, Softmax Entropy and Image Re-synthesis on Lost \& Found test set. 
      }
  \label{fig:resynthesis_visualize_fishy}

 \end{figure*}

Fig.~\ref{fig:resynthesis_visualize_fishy} further visualizes the comparison between JEM and our SLEEG on Lost \& Found test set. Notably, SLEEG can consistently perform better than JEM when detecting objects with various scales and distances. Beside JEM, Additionally, Fig.~\ref{fig:resynthesis_visualize_fishy} and Fig.~\ref{fig:resynthesis_fslaf} describe the comparison of visualization results between the classic softmax entropy~\cite{MSP} and image re-synthesis~\cite{imgr} methods and our SLEEG. It can be observed that these methods perform much worse than our SLEEG and are unable to effectively distinguish in-distribution objects from anomalous ones. And there even exists large background regions that are assigned with higher values than anomalous objects. By contrast, our SLEEG generates anomaly maps where unexpected instances are intensively assigned with higher scores, resulting in much fewer false positive pixels as well.

\section{Limitations}

As shown in the Table 1 of the manuscript and Fig.~\ref{fig:densehybrid_static}, our SLEEG performs inferior to SOTA methods that utilize auxiliary OoD data on the FS Static dataset.
This mainly accounts for that the synthetic anomalous objects in FS Static are similar to the instances in the auxiliary dataset that these methods use, i.e., COCO~\cite{lin2014microsoft} and ADE20K~\cite{zhou2019semantic}. Therefore, training with such OoD samples allows models foresee the anomalous objects, resulting in their good performance. Moreover, training in this manner also results in a large gap between these methods and ours in performance on the FS Lost \& Found and Lost \& Found datasets since anomalous objects in the FS Lost \& Found dataset are more realistic. Finally, though our SLEEG performs inferior on FS Static dataset, since the anomaly segmentation task targets at real-world automatic driving scenarios, the ability of tackling realistic unexpected samples is more essential for our SLEEG. Therefore, we focus on utilizing the inherent spatial context to improve the ability of tackling with the distribution discrepancy of normal and anomalies without requiring extra OoD data.

\bigskip

\bibliography{aaai24}

\begin{thebibliography}{67}
\providecommand{\natexlab}[1]{#1}

\bibitem[{Bevandic et~al.()Bevandic, Kreso, Orsic, and Segvic}]{bevandic19gcpr}
Bevandic, P.; Kreso, I.; Orsic, M.; and Segvic, S. ????
\newblock Simultaneous Semantic Segmentation and Outlier Detection in Presence
  of Domain Shift.
\newblock In \emph{41st {DAGM} German Conference, {DAGM} {GCPR} 2019}.

\bibitem[{Bevandi{\'c} et~al.(2018)Bevandi{\'c}, Kre{\v{s}}o, Or{\v{s}}i{\'c},
  and {\v{S}}egvi{\'c}}]{bevandic2018discriminative}
Bevandi{\'c}, P.; Kre{\v{s}}o, I.; Or{\v{s}}i{\'c}, M.; and {\v{S}}egvi{\'c},
  S. 2018.
\newblock Discriminative out-of-distribution detection for semantic
  segmentation.
\newblock \emph{arXiv preprint arXiv:1808.07703}.

\bibitem[{Bevandi{\'c} et~al.(2019)Bevandi{\'c}, Kre{\v{s}}o, Or{\v{s}}i{\'c},
  and {\v{S}}egvi{\'c}}]{bevandic2019simultaneous}
Bevandi{\'c}, P.; Kre{\v{s}}o, I.; Or{\v{s}}i{\'c}, M.; and {\v{S}}egvi{\'c},
  S. 2019.
\newblock Simultaneous semantic segmentation and outlier detection in presence
  of domain shift.
\newblock In \emph{German conference on pattern recognition}, 33--47. Springer.

\bibitem[{Bevandi{\'c} et~al.(2021)Bevandi{\'c}, Kre{\v{s}}o, Or{\v{s}}i{\'c},
  and {\v{S}}egvi{\'c}}]{Denseoutlier2021}
Bevandi{\'c}, P.; Kre{\v{s}}o, I.; Or{\v{s}}i{\'c}, M.; and {\v{S}}egvi{\'c},
  S. 2021.
\newblock Dense outlier detection and open-set recognition based on training
  with noisy negative images.
\newblock \emph{arXiv preprint arXiv:2101.09193}.

\bibitem[{Bevandic et~al.(2021)Bevandic, Kreso, Orsic, and
  Segvic}]{bevandic21arxiv}
Bevandic, P.; Kreso, I.; Orsic, M.; and Segvic, S. 2021.
\newblock Dense outlier detection and open-set recognition based on training
  with noisy negative images.
\newblock \emph{CoRR}, abs/2101.09193.

\bibitem[{Biase et~al.(2021)Biase, Blum, Siegwart, and Cadena}]{biase21cvpr}
Biase, G.~D.; Blum, H.; Siegwart, R.; and Cadena, C. 2021.
\newblock Pixel-Wise Anomaly Detection in Complex Driving Scenes.
\newblock In \emph{Computer Vision and Pattern Recognition, {CVPR}}.

\bibitem[{Blum et~al.(2019)Blum, Sarlin, Nieto, Siegwart, and
  Cadena}]{blum2019fishyscapes}
Blum, H.; Sarlin, P.-E.; Nieto, J.; Siegwart, R.; and Cadena, C. 2019.
\newblock Fishyscapes: A benchmark for safe semantic segmentation in autonomous
  driving.
\newblock In \emph{Proc. of the IEEE/CVF International Conference on Computer
  Vision Workshops}, 0--0.

\bibitem[{Blum et~al.(2021{\natexlab{a}})Blum, Sarlin, Nieto, Siegwart, and
  Cadena}]{blum2021fishyscapes}
Blum, H.; Sarlin, P.-E.; Nieto, J.; Siegwart, R.; and Cadena, C.
  2021{\natexlab{a}}.
\newblock The fishyscapes benchmark: Measuring blind spots in semantic
  segmentation.
\newblock \emph{International Journal of Computer Vision}, 129(11): 3119--3135.

\bibitem[{Blum et~al.(2021{\natexlab{b}})Blum, Sarlin, Nieto, Siegwart, and
  Cadena}]{blum21ijcv}
Blum, H.; Sarlin, P.-E.; Nieto, J.; Siegwart, R.; and Cadena, C.
  2021{\natexlab{b}}.
\newblock The Fishyscapes Benchmark: Measuring Blind Spots in Semantic
  Segmentation.
\newblock \emph{International Journal of Computer Vision}, 129(11): 3119--3135.

\bibitem[{Bogdoll, Nitsche, and Z{\"o}llner(2022)}]{cvpr2022_survey}
Bogdoll, D.; Nitsche, M.; and Z{\"o}llner, J.~M. 2022.
\newblock Anomaly Detection in Autonomous Driving: A Survey.
\newblock In \emph{Proc. IEEE Conference on Computer Vision and Pattern
  Recognition (CVPR)}, 4488--4499.

\bibitem[{Cen et~al.(2021)Cen, Yun, Cai, Wang, and Liu}]{cen21iccv}
Cen, J.; Yun, P.; Cai, J.; Wang, M.~Y.; and Liu, M. 2021.
\newblock Deep Metric Learning for Open World Semantic Segmentation.
\newblock In \emph{Proceedings of the IEEE/CVF International Conference on
  Computer Vision (ICCV)}, 15333--15342.

\bibitem[{Chan et~al.(2021{\natexlab{a}})Chan, Lis, Uhlemeyer, Blum, Honari,
  Siegwart, Fua, Salzmann, and Rottmann}]{2021SegmentMeIfYouCan}
Chan, R.; Lis, K.; Uhlemeyer, S.; Blum, H.; Honari, S.; Siegwart, R.; Fua, P.;
  Salzmann, M.; and Rottmann, M. 2021{\natexlab{a}}.
\newblock SegmentMeIfYouCan: A Benchmark for Anomaly Segmentation.

\bibitem[{Chan et~al.(2021{\natexlab{b}})Chan, Lis, Uhlemeyer, Blum, Honari,
  Siegwart, Salzmann, Fua, and Rottmann}]{chan21arxiv}
Chan, R.; Lis, K.; Uhlemeyer, S.; Blum, H.; Honari, S.; Siegwart, R.; Salzmann,
  M.; Fua, P.; and Rottmann, M. 2021{\natexlab{b}}.
\newblock SegmentMeIfYouCan: {A} Benchmark for Anomaly Segmentation.
\newblock \emph{CoRR}, abs/2104.14812.

\bibitem[{Chan, Rottmann, and Gottschalk(2021)}]{chan2021entropy}
Chan, R.; Rottmann, M.; and Gottschalk, H. 2021.
\newblock Entropy maximization and meta classification for out-of-distribution
  detection in semantic segmentation.
\newblock In \emph{Proc. IEEE International Conference on Computer Vision
  (ICCV)}, 5128--5137.

\bibitem[{Chen et~al.(2018)Chen, Zhu, Papandreou, Schroff, and
  Adam}]{chen2018encoder}
Chen, L.-C.; Zhu, Y.; Papandreou, G.; Schroff, F.; and Adam, H. 2018.
\newblock Encoder-decoder with atrous separable convolution for semantic image
  segmentation.
\newblock In \emph{Proc. European Conference on Computer Vision (ECCV)},
  801--818.

\bibitem[{Cordts et~al.(2016)Cordts, Omran, Ramos, Rehfeld, Enzweiler,
  Benenson, Franke, Roth, and Schiele}]{cordts2016cityscapes}
Cordts, M.; Omran, M.; Ramos, S.; Rehfeld, T.; Enzweiler, M.; Benenson, R.;
  Franke, U.; Roth, S.; and Schiele, B. 2016.
\newblock The cityscapes dataset for semantic urban scene understanding.
\newblock In \emph{Proc. IEEE Conference on Computer Vision and Pattern
  Recognition (CVPR)}, 3213--3223.

\bibitem[{Creswell et~al.(2018)Creswell, White, Dumoulin, Arulkumaran,
  Sengupta, and Bharath}]{creswell2018generative}
Creswell, A.; White, T.; Dumoulin, V.; Arulkumaran, K.; Sengupta, B.; and
  Bharath, A.~A. 2018.
\newblock Generative adversarial networks: An overview.
\newblock \emph{IEEE signal processing magazine}, 35(1): 53--65.

\bibitem[{Di~Biase et~al.(2021{\natexlab{a}})Di~Biase, Blum, Siegwart, and
  Cadena}]{synboost2021}
Di~Biase, G.; Blum, H.; Siegwart, R.; and Cadena, C. 2021{\natexlab{a}}.
\newblock Pixel-wise Anomaly Detection in Complex Driving Scenes.
\newblock In \emph{Proc. IEEE Conference on Computer Vision and Pattern
  Recognition (CVPR)}, 16918--16927.

\bibitem[{Di~Biase et~al.(2021{\natexlab{b}})Di~Biase, Blum, Siegwart, and
  Cadena}]{di2021pixel}
Di~Biase, G.; Blum, H.; Siegwart, R.; and Cadena, C. 2021{\natexlab{b}}.
\newblock Pixel-wise anomaly detection in complex driving scenes.
\newblock In \emph{Proc. IEEE Conference on Computer Vision and Pattern
  Recognition (CVPR)}, 16918--16927.

\bibitem[{Everingham et~al.(2010)Everingham, Van~Gool, Williams, Winn, and
  Zisserman}]{everingham2010pascal}
Everingham, M.; Van~Gool, L.; Williams, C.~K.; Winn, J.; and Zisserman, A.
  2010.
\newblock The pascal visual object classes (voc) challenge.
\newblock \emph{International journal of computer vision}, 88(2): 303--338.

\bibitem[{Franchi et~al.(2020{\natexlab{a}})Franchi, Bursuc, Aldea, Dubuisson,
  and Bloch}]{franchi20arxiv}
Franchi, G.; Bursuc, A.; Aldea, E.; Dubuisson, S.; and Bloch, I.
  2020{\natexlab{a}}.
\newblock One Versus all for deep Neural Network Incertitude {(OVNNI)}
  quantification.
\newblock \emph{CoRR}, abs/2006.00954.

\bibitem[{Franchi et~al.(2020{\natexlab{b}})Franchi, Bursuc, Aldea, Dubuisson,
  and Bloch}]{franchi20eccv}
Franchi, G.; Bursuc, A.; Aldea, E.; Dubuisson, S.; and Bloch, I.
  2020{\natexlab{b}}.
\newblock {TRADI:} Tracking Deep Neural Network Weight Distributions.
\newblock In \emph{16th European Conference on Computer Vision, {ECCV}}.

\bibitem[{Grathwohl et~al.(2020)Grathwohl, Wang, Jacobsen, Duvenaud, Norouzi,
  and Swersky}]{grathwohl2019your}
Grathwohl, W.; Wang, K.-C.; Jacobsen, J.-H.; Duvenaud, D.; Norouzi, M.; and
  Swersky, K. 2020.
\newblock Your classifier is secretly an energy based model and you should
  treat it like one.
\newblock In \emph{Proc. International Conference on Learning Representations
  (ICLR)}.

\bibitem[{Grci{\'c}, Bevandi{\'c}, and
  {\v{S}}egvi{\'c}(2022)}]{grcic2022densehybrid}
Grci{\'c}, M.; Bevandi{\'c}, P.; and {\v{S}}egvi{\'c}, S. 2022.
\newblock DenseHybrid: Hybrid Anomaly Detection for Dense Open-set Recognition.
\newblock \emph{arXiv preprint arXiv:2207.02606}.

\bibitem[{Grci\'{c}, Bevandi\'{c}, and \v{S}egvi\'{c}()}]{grcic21visapp}
Grci\'{c}, M.; Bevandi\'{c}, P.; and \v{S}egvi\'{c}, S. ????
\newblock Dense Open-set Recognition with Synthetic Outliers Generated by Real
  {NVP}.
\newblock In \emph{16th International Joint Conference on Computer Vision,
  Imaging and Computer Graphics Theory and Applications, {VISIGRAPP} 2021}.

\bibitem[{Gudovskiy, Okuno, and Nakata(2023)}]{gudovskiy2023concurrent}
Gudovskiy, D.; Okuno, T.; and Nakata, Y. 2023.
\newblock Concurrent Misclassification and Out-of-Distribution Detection for
  Semantic Segmentation via Energy-Based Normalizing Flow.
\newblock \emph{arXiv preprint arXiv:2305.09610}.

\bibitem[{Harris, Stephens et~al.(1988)}]{harris1988combined}
Harris, C.; Stephens, M.; et~al. 1988.
\newblock A combined corner and edge detector.
\newblock In \emph{Alvey vision conference}, volume~15, 10--5244. Manchester,
  UK.

\bibitem[{He et~al.(2016)He, Zhang, Ren, and Sun}]{he2016deep}
He, K.; Zhang, X.; Ren, S.; and Sun, J. 2016.
\newblock Deep residual learning for image recognition.
\newblock In \emph{Proc. IEEE Conference on Computer Vision and Pattern
  Recognition (CVPR)}, 770--778.

\bibitem[{Hendrycks et~al.(2019{\natexlab{a}})Hendrycks, Basart, Mazeika,
  Mostajabi, Steinhardt, and Song}]{streethazards2019}
Hendrycks, D.; Basart, S.; Mazeika, M.; Mostajabi, M.; Steinhardt, J.; and
  Song, D. 2019{\natexlab{a}}.
\newblock Scaling out-of-distribution detection for real-world settings.
\newblock \emph{arXiv preprint arXiv:1911.11132}.

\bibitem[{Hendrycks et~al.(2019{\natexlab{b}})Hendrycks, Basart, Mazeika,
  Mostajabi, Steinhardt, and Song}]{2019Scaling}
Hendrycks, D.; Basart, S.; Mazeika, M.; Mostajabi, M.; Steinhardt, J.; and
  Song, D. 2019{\natexlab{b}}.
\newblock Scaling Out-of-Distribution Detection for Real-World Settings.

\bibitem[{Hendrycks et~al.(2019{\natexlab{c}})Hendrycks, Basart, Mazeika,
  Mostajabi, Steinhardt, and Song}]{hendrycks2019scaling}
Hendrycks, D.; Basart, S.; Mazeika, M.; Mostajabi, M.; Steinhardt, J.; and
  Song, D. 2019{\natexlab{c}}.
\newblock Scaling out-of-distribution detection for real-world settings.
\newblock \emph{arXiv preprint arXiv:1911.11132}.

\bibitem[{Hendrycks et~al.(2019{\natexlab{d}})Hendrycks, Basart, Mazeika,
  Mostajabi, Steinhardt, and Song}]{hendrycks19arxiv}
Hendrycks, D.; Basart, S.; Mazeika, M.; Mostajabi, M.; Steinhardt, J.; and
  Song, D. 2019{\natexlab{d}}.
\newblock Scaling out-of-distribution detection for real-world settings.
\newblock \emph{arXiv preprint arXiv:1911.11132}.

\bibitem[{Hendrycks and Gimpel()}]{hendrycks17iclr}
Hendrycks, D.; and Gimpel, K. ????
\newblock A Baseline for Detecting Misclassified and Out-of-Distribution
  Examples in Neural Networks.
\newblock In \emph{5th International Conference on Learning Representations,
  {ICLR} 2017}.

\bibitem[{Hendrycks and Gimpel(2016)}]{hendrycks2016baseline}
Hendrycks, D.; and Gimpel, K. 2016.
\newblock A baseline for detecting misclassified and out-of-distribution
  examples in neural networks.
\newblock \emph{arXiv preprint arXiv:1610.02136}.

\bibitem[{Hendrycks and Gimpel(2017)}]{MSP}
Hendrycks, D.; and Gimpel, K. 2017.
\newblock A Baseline for Detecting Misclassified and Out-of-Distribution
  Examples in Neural Networks.
\newblock \emph{Proc. International Conference on Learning Representations
  (ICLR)}.

\bibitem[{Hendrycks, Mazeika, and Dietterich(2018)}]{hendrycks2018deep}
Hendrycks, D.; Mazeika, M.; and Dietterich, T. 2018.
\newblock Deep anomaly detection with outlier exposure.
\newblock \emph{arXiv preprint arXiv:1812.04606}.

\bibitem[{Hendrycks, Mazeika, and Dietterich()}]{hendrycks19iclr}
Hendrycks, D.; Mazeika, M.; and Dietterich, T.~G. ????
\newblock Deep Anomaly Detection with Outlier Exposure.
\newblock In \emph{7th International Conference on Learning Representations,
  {ICLR} 2019}.

\bibitem[{Huang et~al.(2019)Huang, Yuan, Guo, Zhang, Chen, and
  Wang}]{huang2019isa}
Huang, L.; Yuan, Y.; Guo, J.; Zhang, C.; Chen, X.; and Wang, J. 2019.
\newblock Interlaced Sparse Self-Attention for Semantic Segmentation.
\newblock \emph{arXiv preprint arXiv:1907.12273}.

\bibitem[{Jung et~al.(2021)Jung, Lee, Gwak, Choi, and
  Choo}]{jung2021standardized}
Jung, S.; Lee, J.; Gwak, D.; Choi, S.; and Choo, J. 2021.
\newblock Standardized max logits: A simple yet effective approach for
  identifying unexpected road obstacles in urban-scene segmentation.
\newblock In \emph{Proc. IEEE International Conference on Computer Vision
  (ICCV)}, 15425--15434.

\bibitem[{Kendall and Gal(2017)}]{kendall17nips}
Kendall, A.; and Gal, Y. 2017.
\newblock What Uncertainties Do We Need in Bayesian Deep Learning for Computer
  Vision?
\newblock In \emph{Neural Information Processing Systems}.

\bibitem[{LeCun et~al.(2006)LeCun, Chopra, Hadsell, Ranzato, and
  Huang}]{lecun2006tutorial}
LeCun, Y.; Chopra, S.; Hadsell, R.; Ranzato, M.; and Huang, F. 2006.
\newblock A tutorial on energy-based learning.
\newblock \emph{Predicting structured data}, 1(0).

\bibitem[{Lee et~al.(2017)Lee, Lee, Lee, and Shin}]{lee2017training}
Lee, K.; Lee, H.; Lee, K.; and Shin, J. 2017.
\newblock Training confidence-calibrated classifiers for detecting
  out-of-distribution samples.
\newblock \emph{arXiv preprint arXiv:1711.09325}.

\bibitem[{Lee et~al.(2018{\natexlab{a}})Lee, Lee, Lee, and
  Shin}]{lee2018simple}
Lee, K.; Lee, K.; Lee, H.; and Shin, J. 2018{\natexlab{a}}.
\newblock A simple unified framework for detecting out-of-distribution samples
  and adversarial attacks.
\newblock \emph{Proc. Advances in Neural Information Processing Systems
  (NeurIPS)}, 31.

\bibitem[{Lee et~al.(2018{\natexlab{b}})Lee, Lee, Lee, and Shin}]{lee18nips}
Lee, K.; Lee, K.; Lee, H.; and Shin, J. 2018{\natexlab{b}}.
\newblock A Simple Unified Framework for Detecting Out-of-Distribution Samples
  and Adversarial Attacks.
\newblock In \emph{Neural Information Processing Systems, NeurIPS}.

\bibitem[{Liang et~al.(2022)Liang, Wang, Miao, and Yang}]{liang2022gmmseg}
Liang, C.; Wang, W.; Miao, J.; and Yang, Y. 2022.
\newblock GMMSeg: Gaussian Mixture based Generative Semantic Segmentation
  Models.
\newblock \emph{arXiv preprint arXiv:2210.02025}.

\bibitem[{Liang, Li, and Srikant()}]{liang18iclr}
Liang, S.; Li, Y.; and Srikant, R. ????
\newblock Enhancing The Reliability of Out-of-distribution Image Detection in
  Neural Networks.
\newblock In \emph{6th International Conference on Learning Representations,
  {ICLR} 2018}.

\bibitem[{Lin et~al.(2014)Lin, Maire, Belongie, Hays, Perona, Ramanan,
  Doll{\'a}r, and Zitnick}]{lin2014microsoft}
Lin, T.-Y.; Maire, M.; Belongie, S.; Hays, J.; Perona, P.; Ramanan, D.;
  Doll{\'a}r, P.; and Zitnick, C.~L. 2014.
\newblock Microsoft coco: Common objects in context.
\newblock In \emph{Proc. European Conference on Computer Vision (ECCV)},
  740--755. Springer.

\bibitem[{Lis et~al.(2019{\natexlab{a}})Lis, Nakka, Fua, and Salzmann}]{imgr}
Lis, K.; Nakka, K.; Fua, P.; and Salzmann, M. 2019{\natexlab{a}}.
\newblock Detecting the Unexpected via Image Resynthesis.
\newblock In \emph{Proc. IEEE International Conference on Computer Vision
  (ICCV)}.

\bibitem[{Lis et~al.(2019{\natexlab{b}})Lis, Nakka, Fua, and
  Salzmann}]{lis2019detecting}
Lis, K.; Nakka, K.; Fua, P.; and Salzmann, M. 2019{\natexlab{b}}.
\newblock Detecting the unexpected via image resynthesis.
\newblock In \emph{Proc. IEEE International Conference on Computer Vision
  (ICCV)}, 2152--2161.

\bibitem[{Lis et~al.(2019{\natexlab{c}})Lis, Nakka, Fua, and
  Salzmann}]{lis19iccv}
Lis, K.; Nakka, K.~K.; Fua, P.; and Salzmann, M. 2019{\natexlab{c}}.
\newblock Detecting the Unexpected via Image Resynthesis.
\newblock In \emph{International Conference on Computer Vision, {ICCV}}.

\bibitem[{Liu et~al.(2020{\natexlab{a}})Liu, Wang, Owens, and
  Li}]{liu2020energy}
Liu, W.; Wang, X.; Owens, J.; and Li, Y. 2020{\natexlab{a}}.
\newblock Energy-based out-of-distribution detection.
\newblock \emph{Proc. Advances in Neural Information Processing Systems
  (NeurIPS)}, 33: 21464--21475.

\bibitem[{Liu et~al.(2020{\natexlab{b}})Liu, Wang, Owens, and
  Li}]{liu20neurips}
Liu, W.; Wang, X.; Owens, J.~D.; and Li, Y. 2020{\natexlab{b}}.
\newblock Energy-based Out-of-distribution Detection.
\newblock In \emph{NeurIPS}.

\bibitem[{Liu et~al.(2023)Liu, Ding, Tian, Pang, Belagiannis, Reid, and
  Carneiro}]{liu2023residual}
Liu, Y.; Ding, C.; Tian, Y.; Pang, G.; Belagiannis, V.; Reid, I.; and Carneiro,
  G. 2023.
\newblock Residual Pattern Learning for Pixel-wise Out-of-Distribution
  Detection in Semantic Segmentation.
\newblock arXiv:2211.14512.

\bibitem[{Liu et~al.(2019)Liu, Wang, Liang, Salakhutdinov, Morency, and
  Ueda}]{liu2019deep}
Liu, Z.; Wang, Z.; Liang, P.~P.; Salakhutdinov, R.~R.; Morency, L.-P.; and
  Ueda, M. 2019.
\newblock Deep gamblers: Learning to abstain with portfolio theory.
\newblock \emph{Proc. Advances in Neural Information Processing Systems
  (NeurIPS)}, 32.

\bibitem[{Malinin and Gales(2018)}]{diric2018}
Malinin, A.; and Gales, M.~J. 2018.
\newblock Predictive Uncertainty Estimation via Prior Networks.
\newblock In \emph{Proc. Advances in Neural Information Processing Systems
  (NeurIPS)}.

\bibitem[{Mukhoti and Gal(2018)}]{bayesian2018}
Mukhoti, J.; and Gal, Y. 2018.
\newblock Evaluating bayesian deep learning methods for semantic segmentation.
\newblock \emph{arXiv preprint arXiv:1811.12709}.

\bibitem[{Otsu(1979)}]{otsu1979threshold}
Otsu, N. 1979.
\newblock A threshold selection method from gray-level histograms.
\newblock \emph{IEEE transactions on systems, man, and cybernetics}, 9(1):
  62--66.

\bibitem[{Pinggera et~al.(2016)Pinggera, Ramos, Gehrig, Franke, Rother, and
  Mester}]{lostAndFound2016}
Pinggera, P.; Ramos, S.; Gehrig, S.; Franke, U.; Rother, C.; and Mester, R.
  2016.
\newblock Lost and found: detecting small road hazards for self-driving
  vehicles.
\newblock In \emph{2016 IEEE/RSJ International Conference on Intelligent Robots
  and Systems (IROS)}, 1099--1106. IEEE.

\bibitem[{Shelhamer(2017)}]{shelhamer2017fully}
Shelhamer. 2017.
\newblock Fully convolutional networks for semantic segmentation.
\newblock \emph{IEEE Transactions on Pattern Analysis and Machine Intelligence
  (TPAMI)}, 39(4): 640--651.

\bibitem[{Sun, Guo, and Li(2021)}]{sun21neurips}
Sun, Y.; Guo, C.; and Li, Y. 2021.
\newblock ReAct: Out-of-distribution Detection With Rectified Activations.
\newblock In \emph{NeurIPS}.

\bibitem[{Tian et~al.(2021)Tian, Liu, Pang, Liu, Chen, and
  Carneiro}]{tian2021pixel}
Tian, Y.; Liu, Y.; Pang, G.; Liu, F.; Chen, Y.; and Carneiro, G. 2021.
\newblock Pixel-wise Energy-biased Abstention Learning for Anomaly Segmentation
  on Complex Urban Driving Scenes.
\newblock \emph{arXiv preprint arXiv:2111.12264}.

\bibitem[{Vojir et~al.(2021{\natexlab{a}})Vojir, {\v{S}}ipka, Aljundi,
  Chumerin, Reino, and Matas}]{vojir2021road}
Vojir, T.; {\v{S}}ipka, T.; Aljundi, R.; Chumerin, N.; Reino, D.~O.; and Matas,
  J. 2021{\natexlab{a}}.
\newblock Road anomaly detection by partial image reconstruction with
  segmentation coupling.
\newblock In \emph{Proc. IEEE International Conference on Computer Vision
  (ICCV)}, 15651--15660.

\bibitem[{Vojir et~al.(2021{\natexlab{b}})Vojir, \v{S}ipka, Aljundi, Chumerin,
  Reino, and Matas}]{vojir21iccv}
Vojir, T.; \v{S}ipka, T.; Aljundi, R.; Chumerin, N.; Reino, D.~O.; and Matas,
  J. 2021{\natexlab{b}}.
\newblock Road Anomaly Detection by Partial Image Reconstruction With
  Segmentation Coupling.
\newblock In \emph{International Conference on Computer Vision, {ICCV}}.

\bibitem[{Xia et~al.(2020{\natexlab{a}})Xia, Zhang, Liu, Shen, and
  Yuille}]{xia2020synthesize}
Xia, Y.; Zhang, Y.; Liu, F.; Shen, W.; and Yuille, A.~L. 2020{\natexlab{a}}.
\newblock Synthesize then compare: Detecting failures and anomalies for
  semantic segmentation.
\newblock In \emph{Proc. European Conference on Computer Vision (ECCV)},
  145--161. Springer.

\bibitem[{Xia et~al.(2020{\natexlab{b}})Xia, Zhang, Liu, Shen, and
  Yuille}]{xia20eccv}
Xia, Y.; Zhang, Y.; Liu, F.; Shen, W.; and Yuille, A.~L. 2020{\natexlab{b}}.
\newblock Synthesize Then Compare: Detecting Failures and Anomalies for
  Semantic Segmentation.
\newblock In \emph{16th European Conference on Computer Vision, {ECCV}}.

\bibitem[{Yuan and Wang(2020)}]{YuanCW20}
Yuan, Y.; and Wang, J. 2020.
\newblock Object-Contextual Representations for Semantic Segmentation.

\bibitem[{Zhou et~al.(2019)Zhou, Zhao, Puig, Xiao, Fidler, Barriuso, and
  Torralba}]{zhou2019semantic}
Zhou, B.; Zhao, H.; Puig, X.; Xiao, T.; Fidler, S.; Barriuso, A.; and Torralba,
  A. 2019.
\newblock Semantic understanding of scenes through the ade20k dataset.
\newblock \emph{International Journal of Computer Vision}, 127(3): 302--321.

\end{thebibliography}

\end{document}